\documentclass[12pt]{iopart}

\usepackage{comment,mfirstuc}
\usepackage{graphicx}
\usepackage{grffile}
\usepackage{algorithm}
\usepackage{algpseudocode}
\usepackage{hyperref}
\usepackage{soul}
\usepackage{listings}
\usepackage{multirow}
\usepackage{dcolumn}
\usepackage{bm}
\usepackage{placeins}
\expandafter\let\csname equation*\endcsname\relax
\expandafter\let\csname endequation*\endcsname\relax
\usepackage{amsmath}
\usepackage{amssymb}
\usepackage{mathtools}
\usepackage{subfig}
\usepackage[bottom=1.0in]{geometry}
\usepackage{cite}
\usepackage{lineno}
\RequirePackage{xcolor}
\usepackage[misc]{ifsym}

\usepackage{tcolorbox} 
\usepackage{stfloats} 

\usepackage{siunitx}

\usepackage{colortbl}
\usepackage{array}
\newcolumntype{P}[1]{>{\centering\arraybackslash}p{#1}}
\newcolumntype{M}[1]{>{\centering\arraybackslash}m{#1}}

\hypersetup{ 
    pdfnewwindow=true,      
    colorlinks=true,       
    linkcolor=blue,         
    citecolor=blue,        
    filecolor=blue,      
    urlcolor=blue        
}  

\algblock{Input}{EndInput}
\algnotext{EndInput}
\algblock{Output}{EndOutput}
\algnotext{EndOutput}

\def\be{\begin{eqnarray} &&} 
 
\def\ee{\end{eqnarray}}

\makeatletter
\newcommand{\mainmatter}{%
  \setcounter{footnote}{0}%
  \patchcmd{\@makefntext}{\fnsymbol}{\arabic}{}{}%
  \patchcmd{\@thefnmark}{\fnsymbol}{\arabic}{}{}%
  \def\@makefnmark{\textsuperscript{\arabic{footnote}}}
  \long\def\@makefntext##1{\parindent 1em\noindent
        \hb@xt@1.8em{%
            \hss\@textsuperscript{\normalfont\@thefnmark}}##1}%
}
\makeatother

\newcommand{\addComment}[2]{
  \expandafter\newcommand\csname #1\endcsname[1]{{\bf \color{#2} \capitalisewords{#1}:\,##1}}
  \expandafter\newcommand\csname #1cor\endcsname[2]{{\color{#2} \capitalisewords{#1}:\,\st{##1}{\bf ##2}}}
  \expandafter\newcommand\csname #1color\endcsname{#2}
}
\addComment{cris}{blue} 
\addComment{james}{red}
\addComment{zisis}{brown}

\newcommand{\gluex}{\textsc{GlueX}}

\begin{document}

\providecommand{\keywords}[1]
{
  \small
  \textbf{Keywords:}  {\color{blue}#1
  }
}

\title[\scriptsize{``Flux+Mutability'': A Conditional Generative Approach to One-Class Classification and Anomaly Detection}]{``Flux+Mutability'': \\A Conditional Generative Approach to One-Class Classification and Anomaly Detection} 



\author{C. Fanelli$^{1,2,\ast}$, J. Giroux$^{3,\star}$, Z. Papandreou$^{3,\ddagger}$}

\address{
$^{1}$ Massachusetts Institute of Technology, Cambridge, Massachusetts 02139, USA\\
$^{2}$ The NSF AI Institute for Artificial Intelligence and Fundamental Interactions, Massachusetts 02139, USA,\\
$^{3}$ University of Regina, Regina, SK S4S 0A2, Canada
}

\ead{{\color{blue}
$^{\ast}$cfanelli@mit.edu,
$^{\star}$jng887@uregina.ca, 
$^{\ddagger}$zisis@uregina.ca
} }

\vspace{10pt}
\begin{indented}
\item[]\today
\end{indented}


\begin{abstract}

Anomaly Detection is becoming increasingly popular within the experimental physics community. At experiments such as the Large Hadron Collider, anomaly detection is at the forefront of finding new physics beyond the Standard Model. This paper details the implementation of a novel Machine Learning architecture, called Flux+Mutability, which combines cutting-edge conditional generative models with clustering algorithms. In the `flux' stage we learn the distribution of a reference class. The `mutability' stage at inference addresses if data significantly deviates from the reference class. We demonstrate the validity of our approach and its connection to multiple problems spanning from one-class
classification to anomaly detection. 
In particular, we apply our method to the isolation of neutral showers in an electromagnetic calorimeter and 
show its performance in detecting anomalous dijets events from standard QCD background. 
This approach limits assumptions on the reference sample and remains agnostic to the complementary class of objects of a given problem.
 We describe the possibility of dynamically generating a reference population and defining selection criteria via quantile cuts.
 Remarkably this flexible architecture can be deployed for a wide range of problems, and applications like multi-class classification or data quality control are left for further exploration.


\end{abstract}


%
%
%
%
%

\mainmatter

\section{Introduction}\label{sec:intro}

Nuclear and particle physics are often characterized by problems where one needs to identify particles or events that (i) \textit{belong to} or (ii) \textit{deviate significantly from} a specific `reference' class. 
In the first case we refer to one-class classification (OCC) ---  to identify objects of the reference class amongst all objects, in the latter to anomaly detection (AD) --- which can leverage on OCC to detect abnormal data points compared to the reference class.  

Examples of OCC can be found in an extensive literature review provided by \cite{seliya2021literature}. As for AD, there is a growing number of applications that span from accelerator operations to  physics analyses, 
 the latter being of great interest for example at the Large Hadron Collider (LHC) since new physics beyond Standard Model (BSM) remains elusive (as discussed, \textit{e.g.}, in \cite{nachman2020anomaly, Kasieczka_2021, nachman2020anomaly_density, schwartz_2021}).\footnote{We refer the reader to \cite{nachman2020anomaly} for an introduction as well as to a description of the LHC Olympics challenge in \cite{Kasieczka_2021} and application utilizing this dataset, \textit{e.g.}, \cite{nachman2020anomaly_density, schwartz_2021}.}  
In both cases, one typically deals with multiple features that vary as a function of the phase space of the final state particles reconstructed in the detector.

In this paper we introduce a novel approach to cope with OCC and AD that leverages two different stages that we call ``flux'' and ``mutability'' (F+M).   
In the first stage, flux, we learn the distributions of the reference class data, and in doing that we utilize a combination of conditional autoencoders (cAE) and flow-based models, particularly conditional masked autoregressive flows (cMAF), which are conditioned to the kinematics of the particles or events we are trying to distinguish. As we will describe herein, this allows us to augment the space of the features with a gain in classification performance. 
The second stage, mutability, consists of addressing if the data in the inference phase --- which undergoes a forward pass in the cAE has significantly deviated from the reference class. 
In other words, at a given kinematics one can dynamically generate a reference cluster in the augmented space and measure if an object belongs to the reference cluster or not.\footnote{The term \textit{reference cluster} has been introduced in this work to indicate a large set of points in the augmented (residual plus reconstructed) space representative of a class of particles. Conceptually, this can be considered a probability density function (PDF) in that space.}    
Hierarchical-based clustering (namely, \textit{Hierarchical Density-Based Spatial Clustering of Applications with Noise} (HDBSCAN) \cite{mcinnes2017hdbscan}) is used to fit these data and come up with a probability cut that provides the confidence level for an object to belong or not to the reference class.

In this work we focus on two applications: i) distinguishing neutrons from photons in the Barrel Calorimeter (BCAL) of the \gluex \ experiment \cite{gluex_2021} where neutrons in certain kinematic regions are difficult to simulate or isolate from real data and photons are therefore used as reference class; and ii) identifying possible rare BSM dijet events from QCD dijet background events at LHC \cite{Kasieczka_2021}, the latter representing our reference class. 
One of the advantages of our novel and flexible architecture described in the following sections, is that it relies only on the `reference' class and remains agnostic to the class of objects complementary to the reference class during both the training and inference phases. 
The two classes can be thought of as `signal' and `background' in physics applications. 
When using strictly supervised methods instead, the model typically requires both signal and background as input in order to learn the feature space and produce a binary output. While these algorithms can be efficient and accurate, they are limited by the quality of data we inject. That is to say, they are prone to any bias we may introduce when constructing our training samples. 
This can be critical when our control over one of the two classes, (\textit{e.g.}, the background), is limited and we need to rely on the other class (\textit{e.g.} the signal) or vice versa. 
Using an architecture that relies on the information of one class removes any assumptions we must make about the complementary class and can be extremely important for different applications: for example, it can be utilized for anomaly detection of rare events as well as a method to increase the purity of samples when the original set of real data is characterized by two classes only. 

The remainder of the paper is structured in the following way: In Sec.~\ref{sec:architecture} we will describe the developed architecture and provide a detailed discussion of the training and inference phases. In
Sec.~\ref{sec:applications} we introduce the two problems that our architecture has been applied to: detection of neutrons within the \gluex \ BCAL, and tagging of $Z^{\prime} \rightarrow t \bar{t}$ from QCD dijet background.
In Sec.~\ref{sec:results} we present our analysis and results. 
Finally, in Sec.~\ref{sec:summary} we conclude  with a summary and perspectives on future work.

\section{Flux + Mutability}\label{sec:architecture}

The F+M approach can be broken into three components, 
 namely: (i) a cAE, (ii) a cMAF, and (iii) a clustering algorithm.

(i) The cAE is trained to reconstruct features as a function of kinematic parameters. In this paper we will show two examples: a) identifying single neutral showers that depend on 14 reconstructed observables which vary as a function of the shower energy and location in the BCAL calorimeter at \gluex; and b) analyzing topologies of events at LHC characterized by 2 jets, which are described by 15 reconstructed observables that depend on the transverse momentum of the jets. Note that in both cases the kinematic variables are continuous. 
The reader can find more details about these datasets in Sec. \ref{sec:applications}.

This model is trained first, independently of the cMAF, deploying Huber Loss (see Eq. \ref{eq:huber}):
\begin{equation}\label{eq:huber}
    \mathcal{L}_{cAE} =
    \begin{cases} 	\frac{1}{2}(x - x^{'})^2 , & for \  |x - x^{'}| \leq \delta \; \; \; \; \; \; \; \\ 		
                      \delta (|x - x^{'}| - \frac{1}{2}\delta)  ,    &\text{otherwise}. 		
    \end{cases} 	 
\end{equation}
Using this trained model we forward pass all training samples and obtain both the reconstructed vectors ($\boldsymbol{x^{'}}$) and the residual vectors ($\boldsymbol{x^{'}} - \boldsymbol{x}$) which are then combined into an augmented space. 
Namely, the augmented space will consist of 28 dimensions for the \gluex \ and 30 dimensions for the LHC problems, respectively.

(ii) This augmented dataset is then used to train the cMAF. 
Let $\boldsymbol{x} \in \boldsymbol{X}$ denote an element from the set of input vectors within the training dataset, $\boldsymbol{k} \in \boldsymbol{K}$ the conditional vector for the kinematics, and $\boldsymbol{z} \in \boldsymbol{Z}$ represent the transformed Gaussian vector given by the invertible bijection $f$.\footnote{A bijection is a mathematical function that is both surjective (onto) and injective (one-to-one). Exact details on the bijection $f$ can be found in \cite{maf}.}. 
A conditional flow with N layers can be described by:
\begin{equation}\label{eq:trans}
\boldsymbol{x} = f(\boldsymbol{z}) = f_N \circ f_{N-1} \circ ... f_{1}(\boldsymbol{z_{0}})
\end{equation}
The logarithm of the transformed probability is then given by Eq. \ref{eq:trans_prob}, where $\pi$ denotes the probability under a Gaussian distribution: 
\begin{equation}\label{eq:trans_prob}
    \log p(\boldsymbol{x}|\boldsymbol{k}) = \log \pi (f^{-1}(\boldsymbol{x}) | \boldsymbol{k}) - \sum_{i=1}^{N} \log \left|det \left(\frac{\partial f_{i}^{-1}(\boldsymbol{x})}{\partial \boldsymbol{x}} \right)\right|
\end{equation}
The loss function is then given by the negative log-likelihood:

\begin{equation}\label{eq:cmaf_loss}
    \mathcal{L}_{cMAF} = -\frac{1}{|\boldsymbol{X}|}\sum_{\boldsymbol{x} \in \boldsymbol{X}} \log p(\boldsymbol{x}|\boldsymbol{k})
\end{equation}
It has been found that using the features reconstructed by cAE ($\boldsymbol{x^{'}}$) instead of the original features ($\boldsymbol{x}$) allows for a better separation of classes at inference. 
We also observed that an augmented space made by the recontructed features and the residuals ($\boldsymbol{x^{'}} - \boldsymbol{x}$) increases the separation power. 
As it will be shown in Sec.~\ref{sec:results}, our reference class data at each kinematics can be represented by a cluster in the feature space that is normalized on a hypersphere. 
It turns out that features provide localization in space, while residuals push events deemed ``anomalous'' radially outward, in otherwords, the augmented space with residuals allows the extraction of clusters originally nested within the main population. 
Hence, cMAF is trained on the augmented space of residuals and reconstructed features. 
The conditions remain the kinematic variables discussed prior, although now we must normalize them on the interval (0,1) to allow better convergence of the flow network. Normalization of the conditions for the cAE is not mandatory, although if the conditions exist over a large domain they should be normalized prior to injection.%
 The flow network will be used as the conditional generator to form the \textit{reference cluster} in the augmented space as a function of the kinematics.\footnote{As explained in Sec. \ref{sec:results}, for the \gluex \ BCAL problem the reference cluster corresponds to the photon showers class, whereas for the LHC jet problem it corresponds to the QCD dijet sample.}

(iii) The last part of the architecture consists in clustering based on HDBSCAN. This allows us to fit the objects in the inference phase with respect to the reference cluster on an object-by-object basis, \textit{i.e.}, on a \textit{particle-by-particle} basis for the neutral shower identification problem of \gluex, and on an \textit{event-by-event} basis for the LHC jet problem, as described in Sec. \ref{sec:results}.

The following provides more details on the approach taken to deal with certain aspects that characterize the F+M architecture. 
Sec.~\ref{subsec:continuous_generation} describes the continuous conditional generation; Sec.~\ref{subsec:clustering} covers the separation via clustering and the choice of the dynamic cuts that are applied; finally Sec.~\ref{subsec:inference} provides a global overview of the workflow during the inference phase with Fig. \ref{fig:inference_model} depicting the connection of the components (i), (ii), and (iii) described in this section.

\subsection{Continuous conditional generation}\label{subsec:continuous_generation}

Continuous conditions give rise to the problem of sparsity within the dataset, meaning low numbers of events per condition, \textit{i.e.},in different kinematic domains. The obvious method to circumvent this issue is pre-binning of conditions such that they become discrete. We instead choose to take a different approach, in which we allow conditions to remain continuous but enforce sampling from restricted domains.
This is achieved using Kernel Density Estimation (KDE) to model the transformed probability distribution of the training data in kinematic bins. That is to say, for each bin in the conditional space, we form a density estimation object in which we can call upon to sample sets of latent vectors from restricted domains. %
In inference, we use the inference particle's kinematics to map to a KDE model, fit on the training samples transformed distribution. This model then generates a sample in the $2N$ augmented space of transformed features and residuals, where $N$ is the dimensionality of the feature space.\footnote{`Transformed' referring to the Gaussian space of the invertible bijection of the Normalizing Flow.} We then concatenate the original inference kinematics to each generated $2N$ Gaussian vector, forming a vector of size $2N + L$ and forward pass this vector through the cMAF to generate our augmented space.

\subsection{Separation via clustering and dynamic cut}\label{subsec:clustering}

We introduce an \textit{outlier score} which is obtained comparing each object candidate (\textit{e.g.}, either a particle or a physics event) to the reference cluster. 
This can be thought of as a probability of being an outlier with respect to the reference class. In other words, it corresponds to the complementary probability of being an inlier:

\begin{equation}\label{eq:outlier_score}
    P_{out} = 1 - P_{in}
\end{equation}

This metric is used to make a decision if the object is more likely to belong to the reference class or not.
Every point of the reference cluster is characterized by an inlier probability and therefore by an outlier score. In what follows we detail the process of generating the outlier distribution for the reference cluster. We will then compare the outlier score of the candidate object to the reference distribution.

The following is a brief and concise description of the HDBSCAN algorithm. The reader can find more details about the algorithm in \cite{hdbscan_imp, campello2015hierarchical}; for our implementation in particular we utilized the documentation in \cite{mcinnes2017hdbscan}. 
HDBSCAN utilizes the \textit{mutual reachability} distance between the points to form a weighted graph, which is in turn used to build the clustering hierarchy.\footnote{The \textit{mutual reachability} between two points (x,y) is defined as the maximum value among the \textit{core distance} of x, \textit{the core distance} of y, and the distance (\textit{e.g.}, Euclidean) between x and y. The \textit{core distance} (\textit{e.g.}, the k$_{th}$ nearest neighbor) defines the density of neighbour points. Additional details can be found in \cite{campello2015hierarchical}.
}
Heuristically the density corresponds to the inverse of a distance, $\lambda = \frac{1}{distance}$; the smaller the distance between points the higher the density. 
 In order to get more information on the clustering, a \textit{condensed tree} is formed.
This tree contains clusters and each cluster contains underlying leaf clusters. This stems from the hierarchical nature of the algorithm. %
In the algorithm the user has control over hyperparameters such as ``min\_samples" which sets a lower limit on the number of points needed to be considered a core point and for the algorithm to perform any \textit{mutual reachability} calculation. 
The algorithm prioritizes regions of high density, eventually merging less dense regions with the main cluster if they are reachable under some threshold.
\textit{Persistence} is introduced defining a notion of membership based on how long the point was retained in the cluster, \textit{i.e.}, position in the spanning tree, in order to compare the relative distance scales between a fixed cluster and a point in question.
We also want to consider a density-based notion of membership. This is done via a modification of the Global-Local Outlier Score from Hierarchies (GLOSH)   algorithm \cite{campello2015hierarchical} that allows us to perform the comparison of the points membership persistence with the maximum persistence of the cluster, in order to get a measure of how much of an outlier the point is relative to the fixed cluster.

Combining notions of both distance and density we can now obtain a membership distribution for the reference cluster at each kinematics. This is used to define an outlier metric when classifying new data points. This metric is dynamic in that requires generating a cluster representative of the reference population at any kinematics. 
Therefore one can define a quantile threshold, which can be some outlier score value corresponding to, \textit{e.g.}, keeping 95\% of the population. 

It should be stressed that the quantile cut defines the outlier score of the candidate, that is the probability that an object is an outlier with respect to the reference class.  
Thus, when we classify data via our quantile metric, we define an outlier score cut that corresponds directly to a certain confidence level in data.
This metric allows us to remain completely agnostic with respect to the complimentary class, removing the need for semi-supervised methods---which require an example signal in mind during training, see, \textit{e.g.}, \cite{park2021quasi}---in defining the optimal selection threshold.

\subsection{Workflow at the inference phase}\label{subsec:inference}

\begin{figure}[!]
    \centering
    \includegraphics[trim=0 3cm 0 0,width=0.90\textwidth]{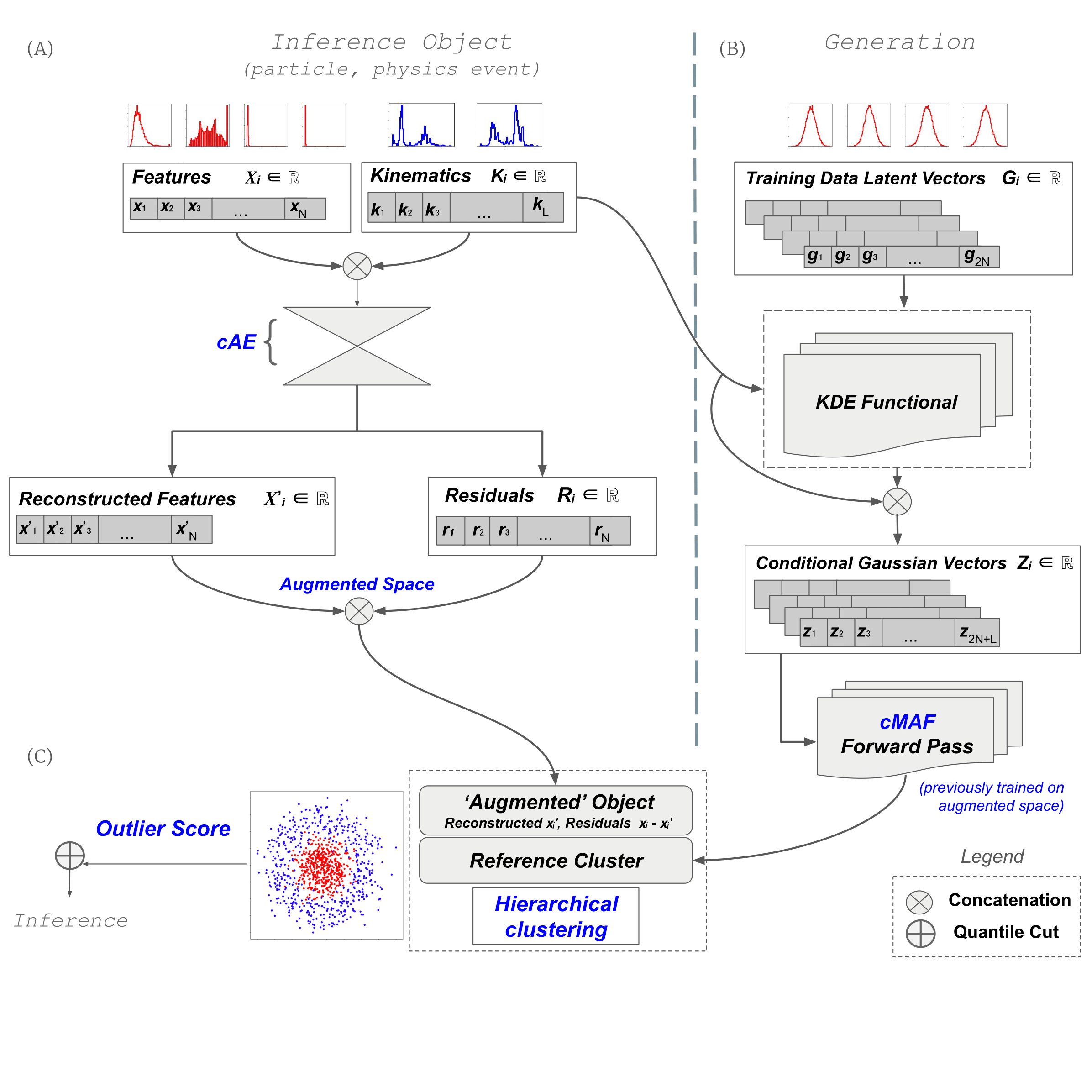}
    \caption{\textbf{Flowchart of the architecture in the inference phase:} 
    The flowchart is described from top to bottom, where the \textit{augmented object} produced by the left column is compared to the \textit{reference cluster} produced by the right column; 
    (A): the inference object is sent through a cAE, in which the conditional learning as a function of the kinematics is done via concatenation. The cAE produces a reconstructed vector used to construct a residual vector ($\boldsymbol{x^{'}} - \boldsymbol{x}$). Reconstructed features and residuals are concatenated as a new augmented space; %
    (B): cMAF is previously trained on the augmented space as a function of continuous conditions. 
    The kinematic vector is mapped to a KDE functional, used to model sub-spaces of the flow-transformed distributions as a function of the kinematics. 
    In the inference phase, the flow network is then fed a sample of Gaussian vectors produced via KDE. %
    The data produced by the flow network is used to form the augmented \textit{reference cluster}. 
    (C): The comparison of the object with the reference cluster produces an outlier score used for classification.}
    \label{fig:inference_model}
\end{figure}

The workflow of F+M is depicted in 
Fig. \ref{fig:inference_model} which describes the flow of an individual object (a shower in the \gluex \ BCAL case or a dijet event in the LHC case) in which we wish to perform inference on. The object is initially fed through the cAE, producing both the reconstructed and residual feature vectors to be augmented.

Subsequently, the kinematics of the input object are mapped to a KDE instance ---pre-fit on the transformed distribution of training samples--- which in turn produces a set of Gaussian vectors from a restricted domain. It is worth mentioning again that using KDE is necessary, given that we are using continuous conditions. For a given condition vector $\boldsymbol{k}$, there are not enough data to reliably sample a full width Gaussian, so we must instead restrict the cMAF sampling domain at the generation stage, for which using KDE is an effective approach. The inference kinematics are concatenated to the vectors and fed to the cMAF for a forward pass. The model is then forced to interpolate over the restricted domain at the generation stage. The generated data is normalized (either on a hypersphere or by applying a standard scaler) and fed to an HDBSCAN clustering instance, forming the reference cluster for the given kinematics. The inference object is added to the cluster, and classification is performed via the dynamic quantile cut. We directly include the inference object in the initial reference cluster since this greatly improves the speed of the algorithm (saving a second clustering). In doing so we are careful to keep the true reference population large such that an individual point has no influence on the quantile metric.

\section{Physics applications and corresponding datasets}\label{sec:applications}

Our approach can be applied to a plenitude of problems in different research areas. 
We selected two examples in particular, one related to the identification of neutral showers caused by neutrons in the \gluex \ BCAL, the other one to BSM dijet events that significantly differ from SM background at LHC. 

\subsection{Neutral showers in the \gluex \ BCAL}\label{subsec:gluex_data}

The \gluex \ experiment, located at Hall~D Jefferson Lab, aims to confirm the predictions of Lattice Quantum Chromodynamics, searching for a class of particles known as exotic hybrid mesons \cite{meyer2015hybrid}.\footnote{\gluex \ is a fixed target photo-production experiment operating at intermediate energies using an electron beam with nominal energy of 12 GeV.}\textsuperscript{,}\footnote{An exotic hybrid meson possesses explicit gluonic degrees of freedom and has quantum numbers predicted by QCD but forbidden by the simple quark model, which includes only quarks and antiquarks.}
The theory predicts multiplets of exotic mesons with different quantum numbers, and the unambiguous establishment of exotic hybrids requires the full mapping of the hybrid multiplet spectrum. This mapping demands the identification of neutral and charged particles in the final state in several topologies and the validation of the results through consistency checks between different decay modes of the same hybrid meson.
Production of charged exotic mesons implies a particle other than a proton must be produced in the reaction. This limits the resulting products to be either a $\Delta$ baryon or a neutron. Charge exchange can also occur in which the proton provides its charge to a positively charged exotic meson, resulting in the production of a neutral $\Delta$ and a neutron. In practice, $\Delta$ baryons are difficult to work with. This is due to large underlying physics background, accompanied by difficulties describing their kinematics, which are necessary for analysis purposes. Ideally, we would like to detect and isolate the neutrons as they do not require detailed modeling of production and decays, and provide constraints to theoretical predictions.

\begin{figure}[!]
\centering
\includegraphics[width=.7\textwidth]{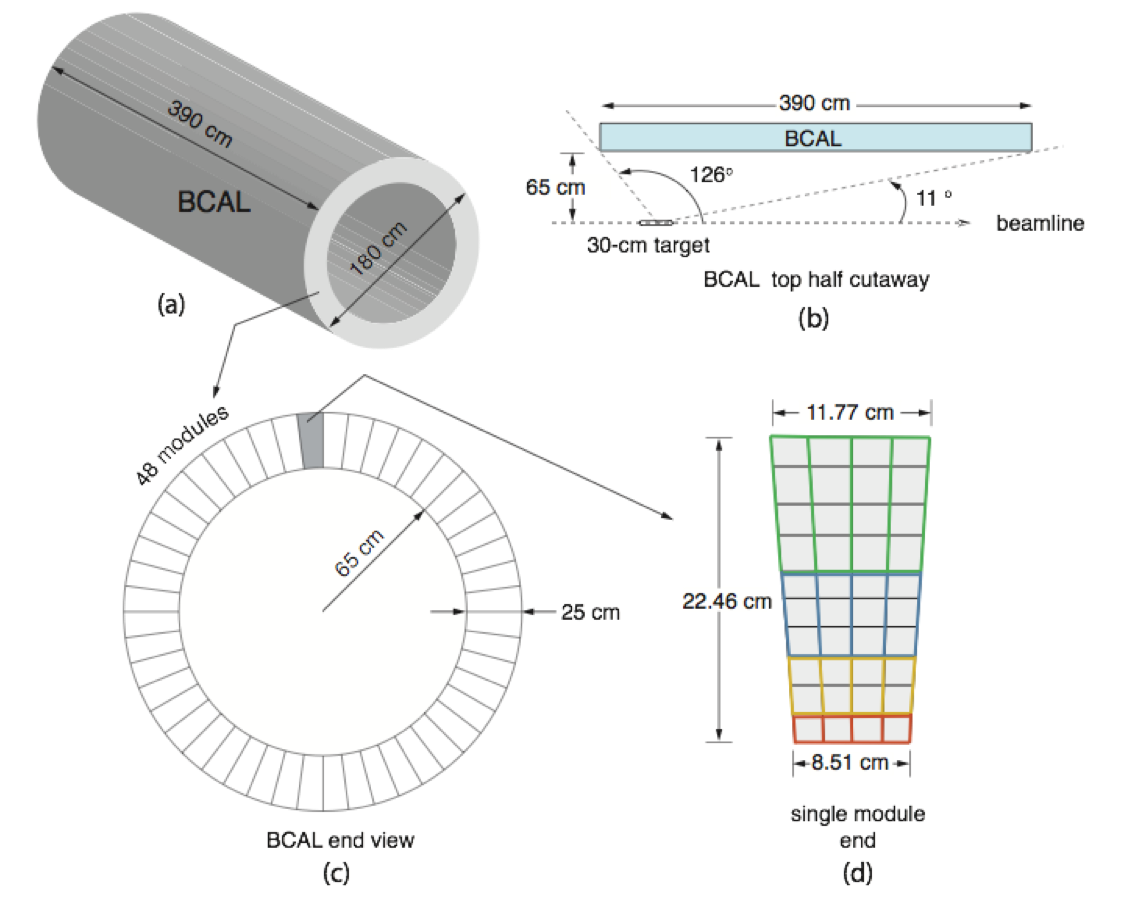}
\caption{\textbf{Sketch of barrel calorimeter readout:} (a) BCAL schematic; (b) a BCAL module side view; (c) end view of the BCAL showing all 48 modules and (d) an end view of a single module showing readout segmentation in four rings (inner to outer) and 16 summed readout zones demarcated by colors. More details can be found in \cite{beattie2018construction}.}
\label{fig:bcal_design}
\end{figure}

 We focus on the Barrel Calorimeter of \gluex \ \cite{beattie2018construction}, a \SI{400}{\cm} long (about $115^{\circ}$ in polar angle coverage) electromagnetic calorimeter designed primarily for photon detection. The detector consists of scintillating fibres compressed between thin layers of lead (see Fig. \ref{fig:bcal_design}).
The detector is segmented into 48 azimuthal modules. 
Each module is partitioned into four readout channels, consisting of double-ended readout using Silicon Photo-Multipliers (SiPM). The \gluex \ photon beam is incident on a liquid hydrogen target ($\gamma + p$) which results in many different final states (termed ``reaction channels"). Many of these particles leave the target and strike the BCAL, creating electromagnetic showers within the detector.  It is from these showers we attempt to classify particles based on their profiles. 
Low-level and high-level observables reconstructed in BCAL are injected in the algorithm as input features (the reader can find a detailed description of the features obtained from the BCAL detector response in \ref{app:features}).

Dealing with neutrons is typically more complicated, in part due to the BCAL detector's response being more difficult to extract compared to a photon. As a result, the extraction of reconstructed neutrons can be affected by sizable uncertainties.\footnote{The agreement between data/simulation in neutron reconstruction is typically $\lesssim$10-15\%, although that discrepancy can be larger for certain final state topologies.} 
Calibrations from real data are also easier for photons, since one can rely on large samples of standard ``candles'' like $\pi^{0}$ decaying into photons which are abundantly produced at \gluex \ and are detected in the BCAL.
With this novel architecture, we are able to limit assumptions we impose on neutrons in the training phase of our algorithm as we rely on half the information of supervised algorithms, namely, the training will be based only on detected photons as they typically provide clear signatures of neutral showers. 
For the proof of principle of our architecture, we will focus on isolated neutral showers, so anything that is not recognized as a photon is then classified as a neutron.\footnote{Isolated neutral showers do not match with a charged track reconstructed by the \gluex \ tracking system, and thus can be clearly separated from charged particle hits.}

The neutral shower reconstruction problem is characterized by 14 features. The dimensionality of this space will be augmented utilizing residuals.
The detector response depends on the kinematics of the particles, that is with which energy and at which position the particle interacts in the calorimeter.\footnote{In the text we equivalently use kinematics or phase space to refer to the kinematic parameters energy and position ($E$, $z$) of the neutral particle.} 
The dependence on the particle kinematics is encoded in our approach through conditional cAE and cMAF, as explained in Sec. \ref{sec:architecture}. 
In fact, as it should clear from Fig. \ref{fig:bcal_design}, the BCAL design has cylindrical geometry, so the two main kinematic parameters that characterize the reconstruction of the neutral shower are the energy $E$ of the particle and the $z$ position at which it strikes the innermost layer. 

 Training and testing data consists of simulated samples, in which generation of only a specific particle type occurs via Geant4 particle gun \cite{AGOSTINELLI2003}, allowing by construction high purity samples in both sets (neutrons and photons).

Particle samples are simulated in such a way to be approximately flat in reconstructed energy $E=$ 200$-$\SI{2200}{\MeV} and in $z=$ 162$-$\SI{262}{\cm} within the BCAL. This corresponds to a region of the phase space that is highly active within the calorimeter.

We initially deploy fiducial cuts on reconstructed widths of showers, namely, width in $z$, radial width and width in time. Doing so removes any artifacts left over from the reconstruction process after simulation, in which are careful to apply loose cuts such that we do not impose restrictions on the data and lose sensitivity to the tails of data distributions. A strict requirement of one shower per event is also required to further eliminate any other interactions (for example split-offs in the photon sample).\footnote{Split-offs are defined as photon conversion to an $e^+ \, e^-$ pair.} 
Since we are interested in only neutrons that mimic photons to a high degree, \textit{i.e.}\ those not easily separated via rectangular cuts, we deploy a tight pre-selection on the radius of a shower (must be within the first 3 BCAL layers) and the amount of energy within the BCAL $4^{th}$ layer (less than \SI{0.1}{\GeV}). The 14 features are comprised of detector response variables and their definitions can be found in \ref{app:features}.
  The entire training dataset consists of $ \approx 1.8$M photons in which we reserve about 10\% for validation.  Testing samples of photons and neutrons are generated independently and each contain about the same number for validation.
 We condition the model on continuous values of $E$ and $z$, modeling the cMAF latent representations in bins of \SI{4}{\cm}, \SI{40}{\MeV}, values that correspond to a coarse representation of the BCAL photon resolution.
 
 The kinematics of the photons and neutrons detected by BCAL are displayed in \ref{app:gluex_kinematics}.
 A comparison of the `original' feature distributions injected in the cAE, the `reconstructed' by the cAE and those dynamically `generated' by cMAF can be found in~\ref{app:gluex_generations}, along with a comparison of the residuals and their corresponding generations.

\subsection{BSM dijet events at LHC}\label{subsec:lhc_data}

Our architecture can be utilized in other problems too.
Despite the multiple searches for physics beyond the Standard Model (BSM) conducted at the LHC, new physics remains elusive as of today.
In the last few years many novel approaches have been developed for AD in order to detect signal events which would stand out as anomalous with respect to a reference background: these span from new unsupervised AD technique leveraging on neural density estimation  \cite{nachman2020anomaly_density} to tag-and-train techniques that can be applied to unlabeled data thus offering to be less sensitive to subtle features of jets which may not be well modeled in simulation \cite{Amram_2021}. 
In this context, our architecture can be utilized to characterize how anomalous an individual event is with respect to a background events by remaining agnostic with respect to the individual events being analyzed. 

In this paper we consider QCD dijet events as background and we look for BSM dijet events from the decay of a $Z^{\prime}$. 
We utilized a suite of jets for SM and BSM particle resonances which is available on Zenodo \cite{BSM_Data}, provided by the authors of \cite{cheng2021variational}.
Primarily, we isolate $Z^{\prime} \rightarrow t \bar{t}$ jets (anomalous signal) from SM QCD dijet (background) in order to remove the varying length feature vectors seen in other BSM datasets, such as $W^{\prime} \rightarrow W + jj$. 
The datasets have been generated with \textsc{MADGRAPH}~\cite{madgraph} and \textsc{PYTHIA8}~\cite{pythia}. The  \textsc{DELPHES}~\cite{delphes} framework has been used for fast detector simulation. 
For a detailed description of the dataset we refer the reader to reference \cite{BSM_Data}.
The simulated QCD \cite{SM_Data} and BSM dijets \cite{BSM_Data} are produced with the same selection criteria.
Clustering of the jets was done using \textsc{FASTJET}~\cite{fastjet}, deploying the anti-$k_T$ algorithm~\cite{anti_kt} with a cone size of $R=1.0$. As stated in~\cite{SM_Data,BSM_Data}, events within the datasets must meet the requirement of the leading jet having transverse momenta $\boldsymbol{p_T}$ $ >$ \SI{450}{\GeV} and the sub-leading jet having $\boldsymbol{p_T} >$ \SI{200}{\GeV}. %

In this case we use only a single conditional, namely the leading jet transverse momentum, and form a fixed length feature vector consisting of the remaining 4 vector properties of the leading jet, its n-subjettiness variables, the sub-leading jet 4 vector and its n-subjettiness variables.\footnote{See \cite{Kasieczka_2020} for mathematical definition of n-subjettiness.} This feature vector then gets augmented with its residual vector from the cAE, resulting in a vector of 30 features at inference including residuals.
We apply a further condition on the datasets, requiring the leading jet to have $\boldsymbol{p_T} <$ \SI{800}{\GeV} in order to provide sufficient data as a function of the conditional parameter. We model the cMAF's transformed space in bins of \SI{1}{\GeV}. 

The architecture is trained on $\approx 600$k QCD dijet events and validated on $\approx 50$k, retaining around $50$k for testing. We use only a single top jet file from \cite{BSM_Data} ($m_t =$ \SI{174}{\GeV}), providing $50$k anomalous events.
More details on the feature distributions of both classes can be found in \ref{app:lhc_generations} which includes also a comparison of the `original' feature distributions injected in the cAE, the `reconstructed' by the cAE and those dynamically `generated' by cMAF along with the residuals and their corresponding generations.

\section{Analysis and results}\label{sec:results}

In what follows, we deploy our architecture on the two different physics scenarios introduced in Sec. \ref{sec:applications}. 

\subsection{Neutral showers classification with the \gluex \ BCAL}\label{subsec:problem1}

We demonstrate the potential of the model as an OCC method for \gluex \ photons, which in turn allows to tag neutron candidates from the sample of isolated neutral showers described in Sec. \ref{subsec:gluex_data}. 
As already explained, there are specific regions in the phase space of the BCAL where simulating the detector response to neutrons is challenging because it is characterized by large uncertainty.%

Our strategy aims at isolating neutrons by applying cuts on the photon showers, the latter taken as the reference class. 
As described in Sec. \ref{sec:architecture}, our approach is unsupervised and agnostic to the neutrons; it allows us to dynamically generate a reference cluster in the augmented space of the features as a function of the particle kinematics.
The reference cluster is used to establish if a new particle is more likely to belong to the photon class or the neutron class. 
A quantile cut is applied on an outlier score to determine the probability of a particle to be a member of the cluster or to be an outlier. 
This approach can be useful when the uncertainty on the distributions of the complementary class (neutrons) is expected to be large compared to the reference class (photons), and the distributions of neutrons and photons cannot be easily separated by standard rectangular selections. 
In such scenarios, fully supervised approaches become less reliable without a proper assessment of the uncertainty quantification.  
Our approach allows to select the true positive rate (TPR) of the reference class which, by construction, is consistent with the quantile cut on the outlier score chosen for the selection. 
%
%
The outlier score of each particle corresponds to a probability of not belonging to the photon reference class, according to Eq.~\eqref{eq:outlier_score}; given that we work with isolated neutral showers which can only be either a photon or a neutron, the outlier score is interpreted as how confident we are to have identified a neutron.

\begin{figure}[!]
    \centering
    \includegraphics[width=0.5\textwidth]{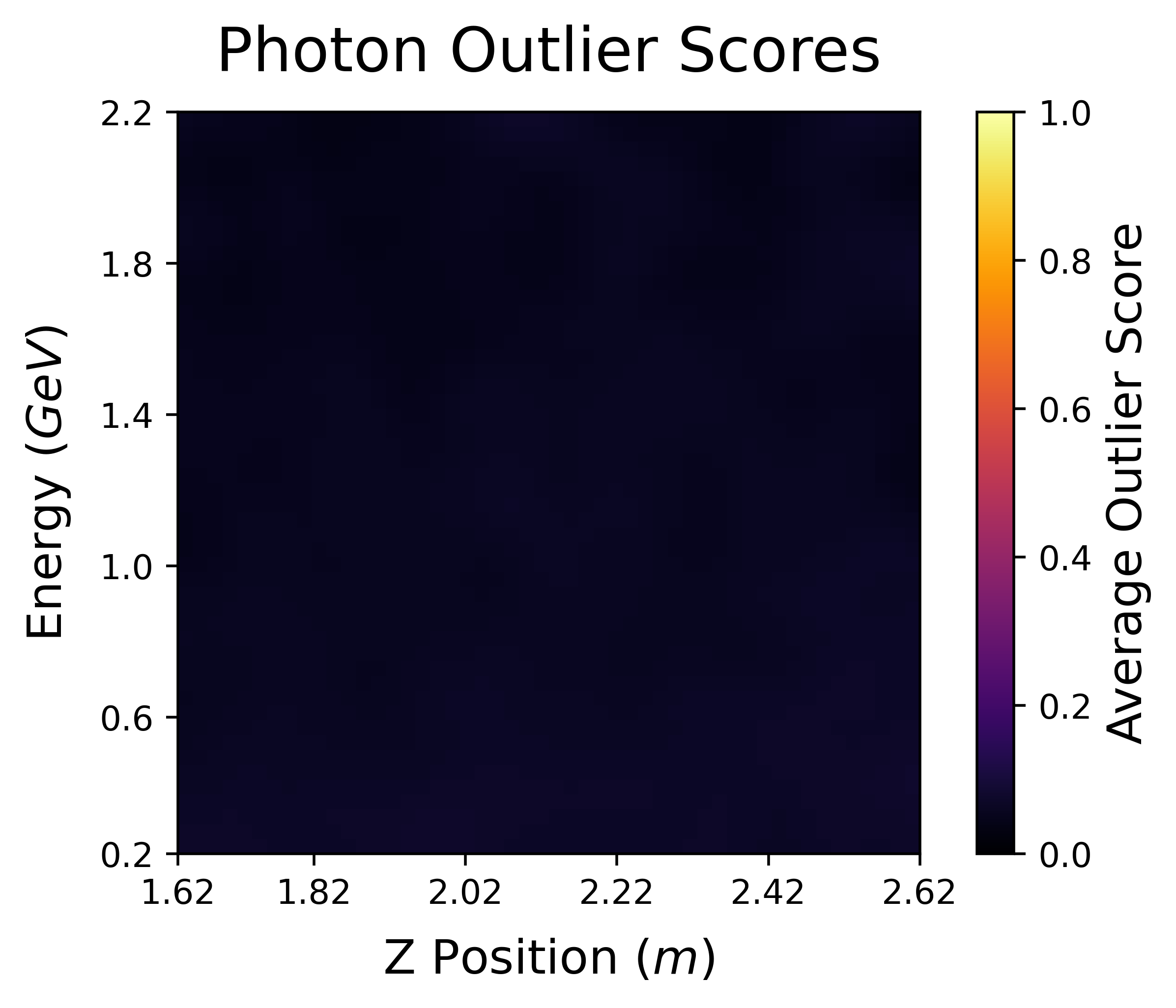}%
    \includegraphics[width=0.5\textwidth]{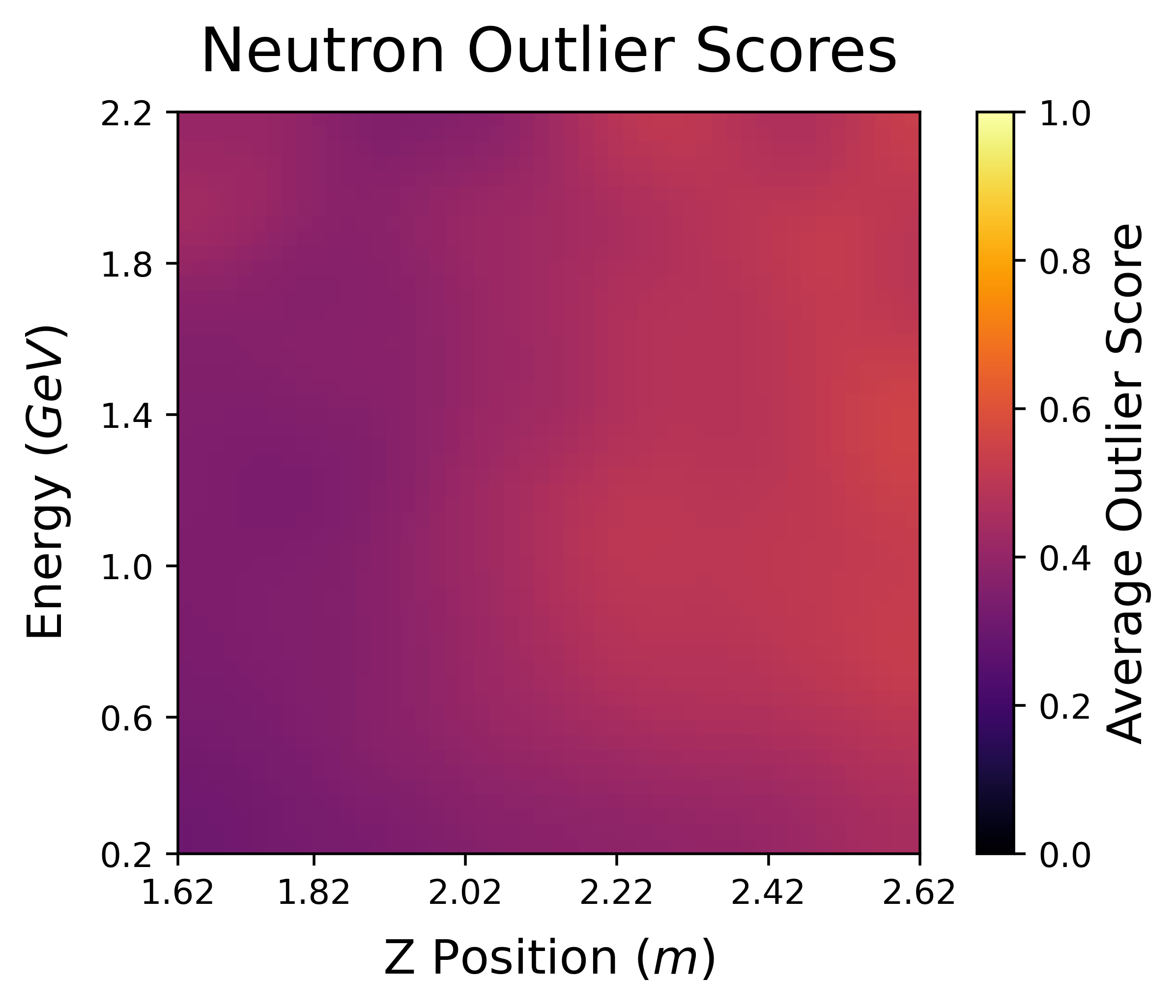}
    \caption{\textbf{Outlier scores:} Average outlier scores (probability of not belonging to the photon class) in bins of \SI{40}{\MeV} and \SI{2}{\cm}. Signal particles (photons, left) display much lower outlier scores (color) on average than background (neutrons, right) across the grid of the phase space (x axis: Z position, y axis: Energy). There are no apparent kinematic dependencies on separation power.
    }
    \label{fig:Outliers_Unsupervised}
\end{figure}

 Fig.~\ref{fig:Outliers_Unsupervised} depicts the average value of the outlier score in bins of $E$ and $z$.\footnote{The binning in $z$ is smaller than our KDE functionals in order to obtain a square grid for plotting.} 
  It is easily seen that the average outlier scores for neutrons are much higher across the entire phase space when comparing both plots.
It is also apparent that the outlier score of photons is flat and close to zero as a function of the kinematics and for neutrons it is large in value and rather uniform in distribution too, despite the reconstructed features do largely depend on the kinematics of the particles (as displayed in Figs. \ref{fig:photons_f_z}, \ref{fig:neutrons_f(z)}). 
 This means the architecture has provided an approximately uniform and good separation power in the phase space we are covering. 
 Deploying a 95\% quantile cut, we obtain a True Positive Rate (TPR) for photons, and a True Negative Rate (TNR) for neutrons of \textbf{95.09\%} and \textbf{52.40\%}, respectively, as summarized in Table~\ref{tab:augmented_comp}. 
 TPR and TNR are quite large and to our knowledge exceed by far results obtained with traditional rectangular selections \cite{beattie_2022}.
 
 We note that by assuming both photon and neutron training datasets are reliable, then deploying a fully supervised model like XGBoost~\cite{xgb} would result in a TPR and TNR at about $92\%$ each, but again this is not the scenario we are tackling here.\footnote{For completeness, we also utilized our F+M approach considering both classes as hypotheses to build a figure of merit of the form of a $\Delta \log{\mathcal{L}}$ to apply a cut on. Results are outperformed by other methods, \textit{e.g.}, XGBoost, which seem more suitable for a fully supervised task.} 
 Efforts have been made to improve the prediction head of the algorithm, in which we replaced the clusterer with a One-Class Support Vector Machine (OC-SVM). The conditional generations (see Fig. \ref{fig:gluex_generations} for an example of generations produced via the cMAF) are then used to fit the OC-SVM and predict outliers (\textit{i.e.}, neutrons). In smaller scale tests it was found that the TPR of photons suffered drastically, at around $56\%$, while offering a slight increase in the TNR of neutrons at around $90\%$. These results are most comparable to the results obtained deploying a $68\%$ quantile cut with clustering, yet pose no real performance increases. Using the OC-SVM also limits our ability to employ a quantile cut.
 The SVM-based method was deemed inferior to the clustering method used in our approach and not explored further.

\begin{table}
\centering
\begin{tabular}{|c|c|c| } 
 \hline
Quantile & TPR & TNR \\ 
 \hline
1$\sigma$ (68\%) & \textbf{68.28 $\pm$ 0.18 \%} & \textbf{87.44 $\pm$ 0.13\%} \\  
 \hline 
2$\sigma$ (95\%) & \textbf{95.09 $\pm$ 0.08} \% & \textbf{52.40 $\pm$ 0.19\%} \\  
 \hline
3$\sigma$ (99\%) & \textbf{98.97 $\pm$ 0.04 \%} & \textbf{34.95 $\pm$ 0.18\%} \\  
 \hline

\end{tabular}
 \caption{\textbf{Summary of \gluex \ BCAL results:} photons are used as the reference class, and results are obtained using different quantile cuts on the outlier score. Fluctuations in TPR are within the error.}
  \label{tab:augmented_comp}
\end{table}

\subsection{Uncertainty in the neutron sample}

We further showcase the utility of our architecture when large uncertainty affects the class complementary to the reference class. 
In our case, the complementary class is that of the neutron candidates, and we know that neutron simulations in the phase space of interest described in this study can differ significantly from real data. 
We show how this affects a fully supervised approach such as XGBoost and make a comparison to our OCC approach which benefits from the usage of residuals.

While the original injected neutron distributions overlap to a great extent with the photon distributions to begin with, we push this overlap to an extreme case: we artificially create an extra testing dataset of neutrons in which we `scale' the neutron features in such a way to make them highly resemble photons. 
We pretend this new sample to be the \textit{actual} data observed in an experiment representative of the true detector response, while we consider the original sample to be the \textit{simulated} data in which we assume there exists $\sim 10-15\%$ difference from actual data. 
Fig. \ref{fig:pert_original} shows a comparison between photons and neutrons of the injected feature distributions. For the neutrons, we also show the case of the `scaled' (\textit{i.e.}, actual) distributions. 
Similarly, Fig.~\ref{fig:pert_recon} shows a comparison  for the distributions of the features as reconstructed by the cAE; Fig.~\ref{fig:pert_residuals} shows the same comparison for the residuals. Finally, Fig. \ref{fig:perturbation_outlier_scores} shows a comparison for the distribution of the outlier scores.

XGBoost is then trained on the simulated neutron sample, and we compare performance on both \textit{simulated} and actual (scaled) samples in the inference phase. 
In this study we use the TPR obtained with XGBoost to define the quantile cut for our architecture and make a comparison. 
Table \ref{tab:perturbation} shows the results of this study. 
XGBoost TNR performance drops from about 92\% to 79\% when deployed to what we consider the actual data in this example. This is the result of a decision boundary obtained in the training phase using an inaccurate neutron sample. 
\begin{table}[!]
\centering
\scalebox{0.9}{
\begin{tabular}{ c c c c c } 
& \multicolumn{2}{c}{Simulation} & \multicolumn{2}{c}{``Actual'' Detector Response} \\
 \hline
 \textbf{Algorithm} & \textbf{TPR} & \textbf{TNR} & \textbf{TPR} & \textbf{TNR} \\ 
 \hline
 XGBoost & $92.15 \pm 0.10 $\%& $91.93 \pm 0.10 $\%&  $92.15 \pm 0.10 $\% & $\textbf{78.82} \pm 0.15$\% \\ 
 \hline
 F + M (Augmented) & $92.28 \pm 0.10 $\%& $60.29 \pm 0.18 $\%&  $92.33 \pm 0.10 $\% & $\textbf{82.71} \pm 0.14$\% \\ 
 \hline
 F + M (Features) & $92.34 \pm 0.10 $\%& $56.14 \pm 0.19 $\%&  $92.34 \pm 0.10 $\% & $50.30 \pm 0.19$\% \\ 
 \hline
\end{tabular}
}
 \caption{\textbf{Neutron sample study:} The table shows a comparison of the performance obtained using two neutron samples, one assumed to be simulated and the other one considered as the ``actual'' detector response to neutrons. Differently from neutrons, photons simulations are more accurate and in agreement with real data. 
 TPR is the true positive rate for photons and TNR the true negative rate for neutrons. XGBoost is trained on simulated photons and neutrons. F+M relies only on simulated photons. XGBoost TNR performance drops when deployed on actual data. The increased TNR of F+M depends on the residual space produced by the cAE which captures a different kinematic dependence of the neutron features compared to that of the photons in the actual case.}
  \label{tab:perturbation}
\end{table}
On the other hand, F+M is an OCC approach that relies on photons only and is agnostic to the neutron sample which is `seen' only during the inference phase. 
The TNR performance increases from 60\% to 83\% when deployed on the ``actual'' neutron sample. 
The reason for the increase is likely due to learning correlations in the kinematics in the augmented space. In fact, while the new neutron features are artificially made more photon-like by scaling their distributions and hence look harder to separate from photons, their correlations with kinematics which cAE is able to pick up changed, and the resulting residuals are more easy to separate. This is further confirmed via the results obtained using the feature space only, in which performance drops in with respect the the unperturbed neutron sample.

A result provided by XGBoost in this example would provide a large TNR of 92\% but it is affected by a large systematic as indeed the true performance would be 79\%. 
F+M training does not depend on the neutron sample, and therefore the TNR performance is not affected by the same large uncertainty and it actually provides in this particular case a larger value of 82\%.
We note that:
\begin{itemize}
    \item The result of the TNR for F+M depends on how the actual data look like in the inference phase, \textit{i.e.}, the opposite result applies by switching labels in Table \ref{tab:perturbation};
    \item OCC is agnostic to the complementary class, which is unlabeled. External physical information can be used to label neutron data, \textit{e.g.}, this may depend on the event topology containing the neutron; 
    \item The outlier score of each particle represents a proxy for the confidence level of its classification; it's worth reminding that the quantile cut is defined by the photon sample and is dynamical in that depends on the kinematics; we notice that the neutron's outlier score increased on average for the new distributions, meaning that the uncertainty on each individual classified neutron is also decreased. More details can be found in Fig.~\ref{fig:perturbation_outlier_scores}. 
\end{itemize}

\subsection{Anomaly detection of dijet events at LHC}\label{subsec:problem2}

We deploy our algorithm for anomaly detection of BSM dijets with respect to a background of QCD dijet events.
We consider the datasets described in Sec. \ref{sec:applications}.
The performance of our algorithm is compared to other works \cite{schwartz_2021, cheng2021variational} that used the same dataset and considered as a metric the area under the receiver operating characteristic curve (AUC). 
%

 However, it should be noted that the AUC by construction is not agnostic to the BSM signal in the anomaly detection problem: in fact it provides model performance as a function of threshold cuts. In order to pick the optimal threshold one must have prior knowledge of the anomalous class itself, that is not always possible and if it is, implies a bias towards the model which in turn becomes weakly supervised.
 There are of course other methods to identify a suitable cut remaining agnostic towards the anomalous sample, yet these values may be far from optimal. Thus AUC can be an inflation of true model performance.

The AUC is obtained by fitting our ROC curve. For each quantile between (0.02,0.98) in steps of 0.02, we compute the TPR and FPR on a random sample of size of 1k (approximately 50/50 of each class), propagating the uncertainty in both efficiencies. 
We report the AUC values in Table.~\ref{tab:auc} and compare to the best values obtained by \cite{schwartz_2021} and \cite{cheng2021variational} using the same dataset. 
In \cite{schwartz_2021, cheng2021variational} different AUC scores are listed based on different settings, loss functions, etc. Some knowledge of the anomalous class is utilized in order to define the optimal threshold.
The AUC score of our architecture is slightly larger than in \cite{schwartz_2021} and consistent within the uncertainty with that of \cite{cheng2021variational}.
It should be noticed that our architecture can be further optimized by tuning its hyperparameters. This will possibly further improve the results shown in Table \ref{tab:auc}. 

\begin{table}[h!]
\centering
\begin{tabular}{|c|c|c|c| } 
 \hline
& Ours & From \cite{schwartz_2021} & From \cite{cheng2021variational}\\ 
 \hline
AUC & $\mathbf{0.885 \pm 0.003}$ & 0.87 & 0.89\\ 
\hline
\end{tabular}
 \caption{\textbf{AUC score comparison:} AUC score comparison between our architecture and two methods, Fraser et al. \cite{schwartz_2021} and Cheng et al. \cite{cheng2021variational}. Our architecture performs on par with \cite{cheng2021variational} within the uncertainty and slightly better than \cite{schwartz_2021}. It should be noticed that our architecture can be further optimized by tuning its hyperparameters.  This will possibly further improve the results.}
  \label{tab:auc}
\end{table}

\begin{figure}[!]
     \centering
     \includegraphics[width=0.49\textwidth]{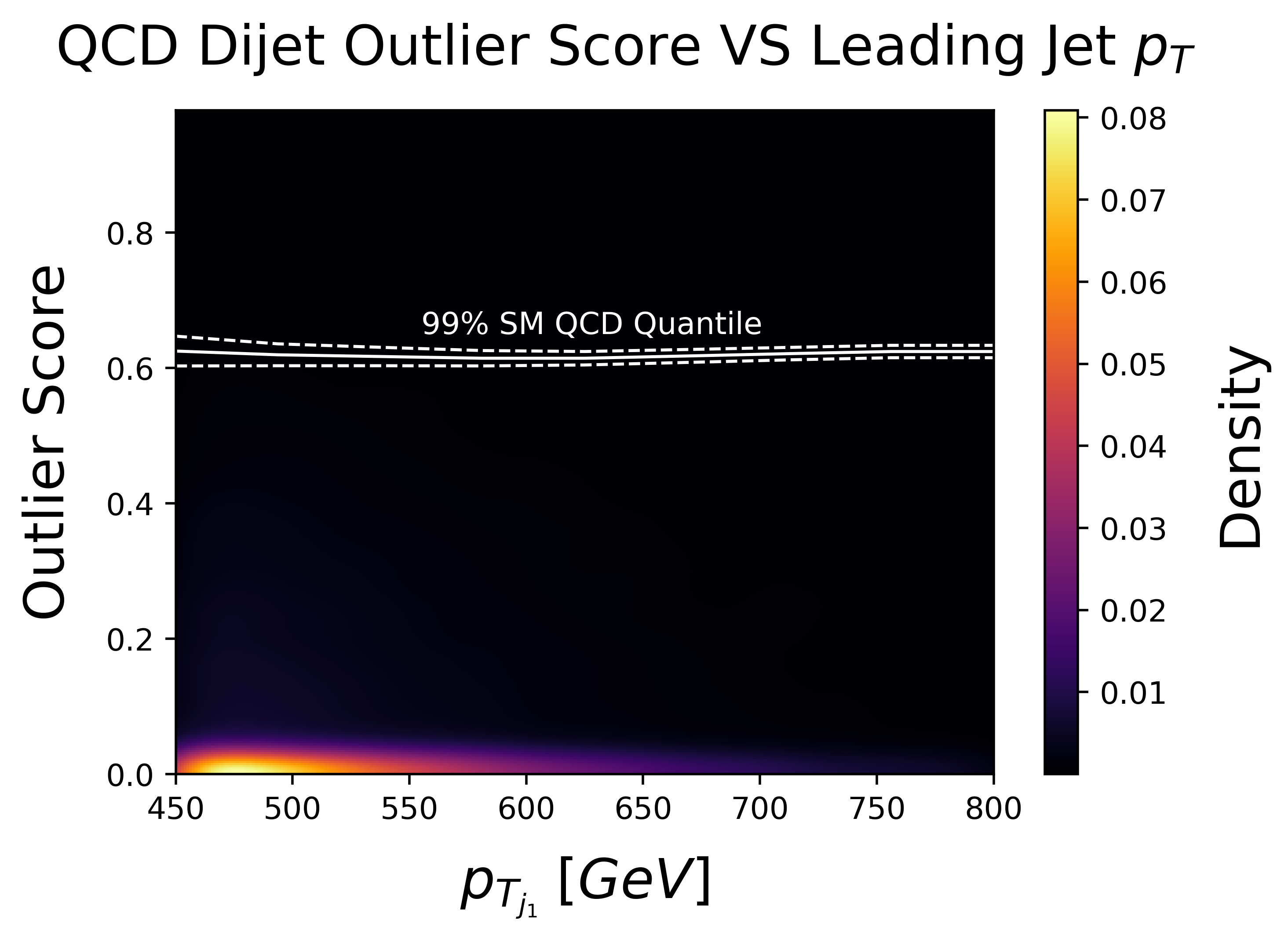} %
     \includegraphics[width=0.49\textwidth]{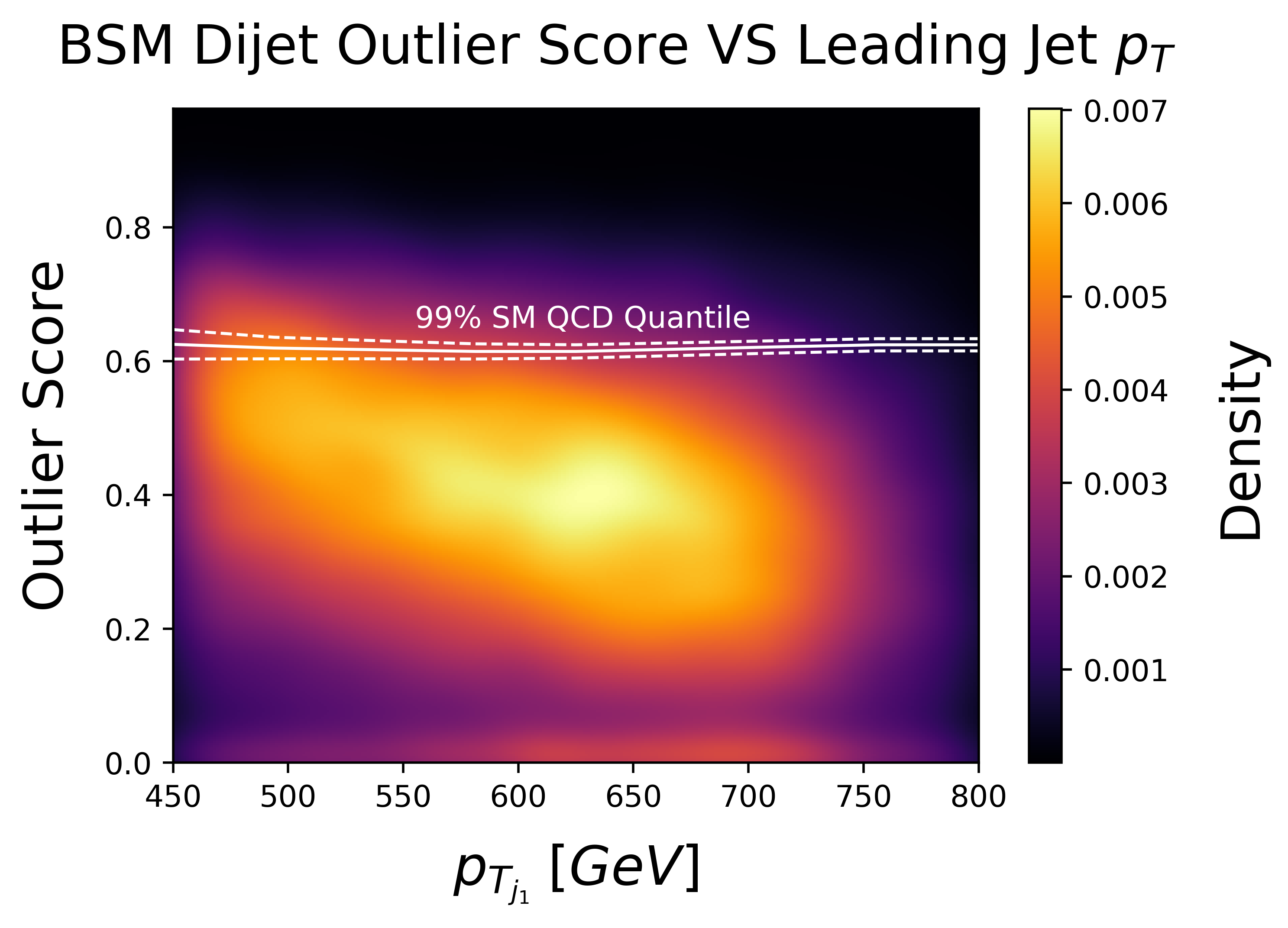}
     \caption{\textbf{Outlier scores as a function of leading jet transverse momentum:} Outlier score as a function of leading jet transverse momentum for QCD dijets (left), and BSM dijets (right) at LHC. The 99\% quantile threshold is overlayed. Opting for large values of the quantile results in isolating the tails of the complementary class distribution if features overlap to a high degree.
     The 1$\sigma$ band has been calculated from reference clusters with 1.5k generated objects.
     }
     \label{fig:LHC_Outliers}
 \end{figure}

In the following we calculate TPR and TNR. The idea is that of remaining agnostic to a potential BSM dijet signal, and changing the quantile cut on the QCD dijet background. 
 For example, one may consider setting the threshold of our architecture to $\approx$ 100\% for QCD dijets, very naively letting only largely anomalous BSM samples to stand out from the selected distributions. 
 For completeness we show different scenarios (including also low quantile cuts) in Table \ref{tab:augmented_comp_lhc}.
  Fig. \ref{fig:LHC_Outliers} depicts the outlier scores of both QCD and BSM dijets at LHC as a function of the leading jet transverse momentum, in which the isolation of the tail of the BSM distribution is visualizable with respect to the 99\% quantile threshold. 
  One may be temped to instead opt for a global cut within the outlier space, however this explicitly demands prior knowledge of the complementary class, directly violating the conditions of anomaly detection frameworks.

\begin{table}
\centering
\begin{tabular}{|c|c|c| } 
 \hline
Quantile & TPR & TNR \\ 
 \hline
 1$\sigma$ (68\%) & \textbf{68.18 $\pm$ 0.22 \%} & \textbf{93.20 $\pm$ 0.06\%} \\  
 \hline 
 2$\sigma$ (95\%) & \textbf{95.15 $\pm$ 0.10 \%} & \textbf{42.40 $\pm$ 0.22\%} \\  
 \hline
 3$\sigma$ (99\%) & \textbf{99.03 $\pm$ 0.05 \%} & \textbf{11.82 $\pm$ 0.14\%} \\  
  \hline
  Fiducial cuts (99\%) & \textbf{98.92 $\pm$ 0.05 \%} & \textbf{2.35 $\pm$ 0.06\%} \\
 \hline
 
 \end{tabular}
 \caption{\textbf{Summary of dijet results at LHC:} results are obtained using different quantile cuts on the outlier score. Fluctuations in TPR are within the error. For comparison, we also include a baseline rectangular selection based on loose fiducial cuts on each feature, defined in such a way to select when combined 99\% of the SM QCD dijet events. 
 }
  \label{tab:augmented_comp_lhc}
\end{table}

As is clear in all our tables the TPR is consistent by construction with the quantile cut applied to the outlier score obtained using the generated reference cluster. 

\subsection{Benefits of the augmented space: residuals}\label{subsec:performance}

We have demonstrated the good performance of our architecture for different physics problems. 
 In order to illustrate the performance increase via residuals, we follow the same inference procedures discussed before except we remove the residuals as input to the cluster. Table.~\ref{tab:GlueX_CERN_Residuals} shows a comparison between results obtained using the feature space and the augmented space of features plus residuals.
The comparison is done for both problems (neutron identification in \gluex \ BCAL and dijet anomaly detection at LHC). 
In both cases, we observe a systematic improvement in the TNR by using the augmented space. 

\begin{figure}[!]
    \centering
    \includegraphics[width=0.31\textwidth]{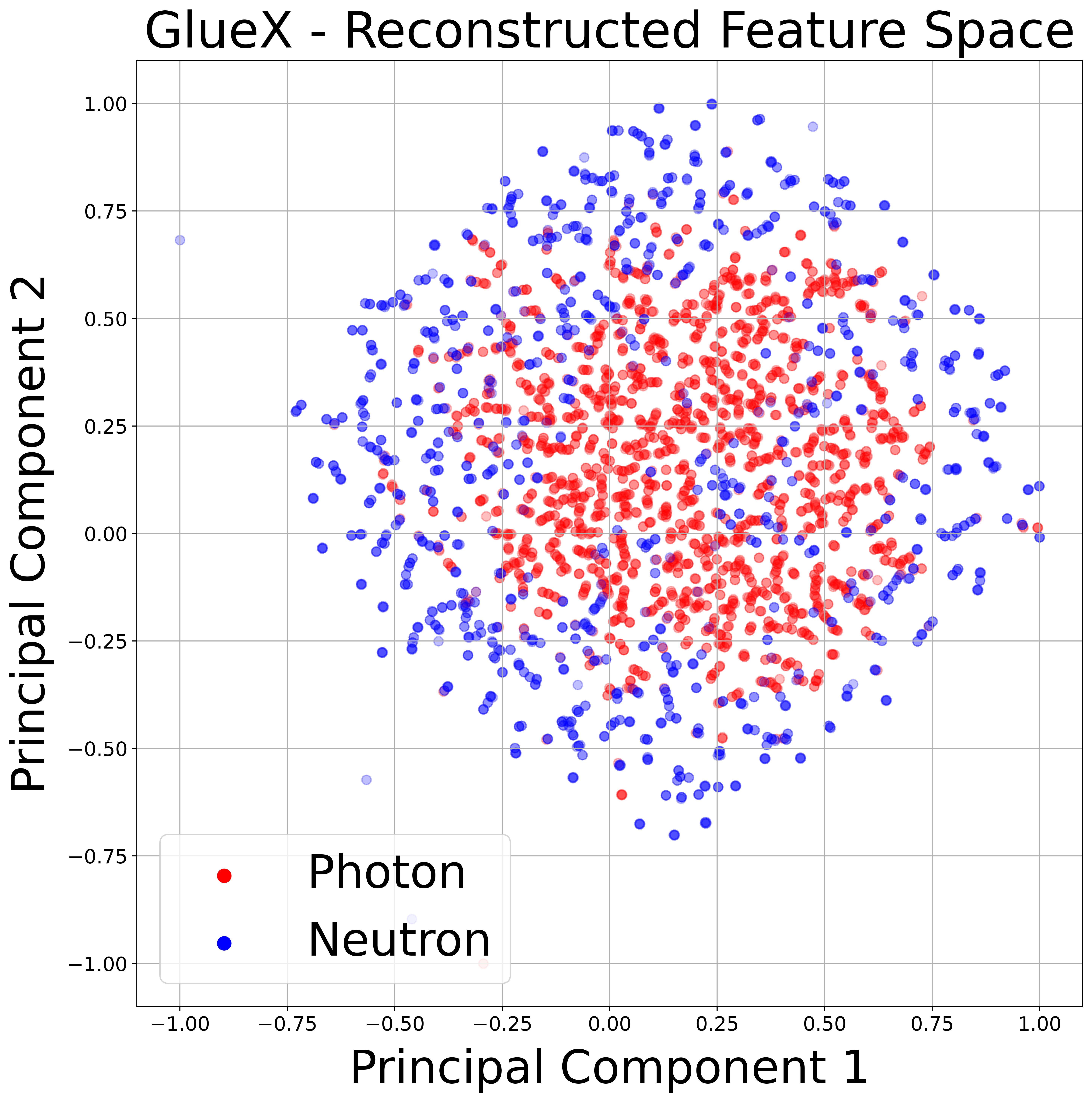}%
    \includegraphics[width=0.31\textwidth]{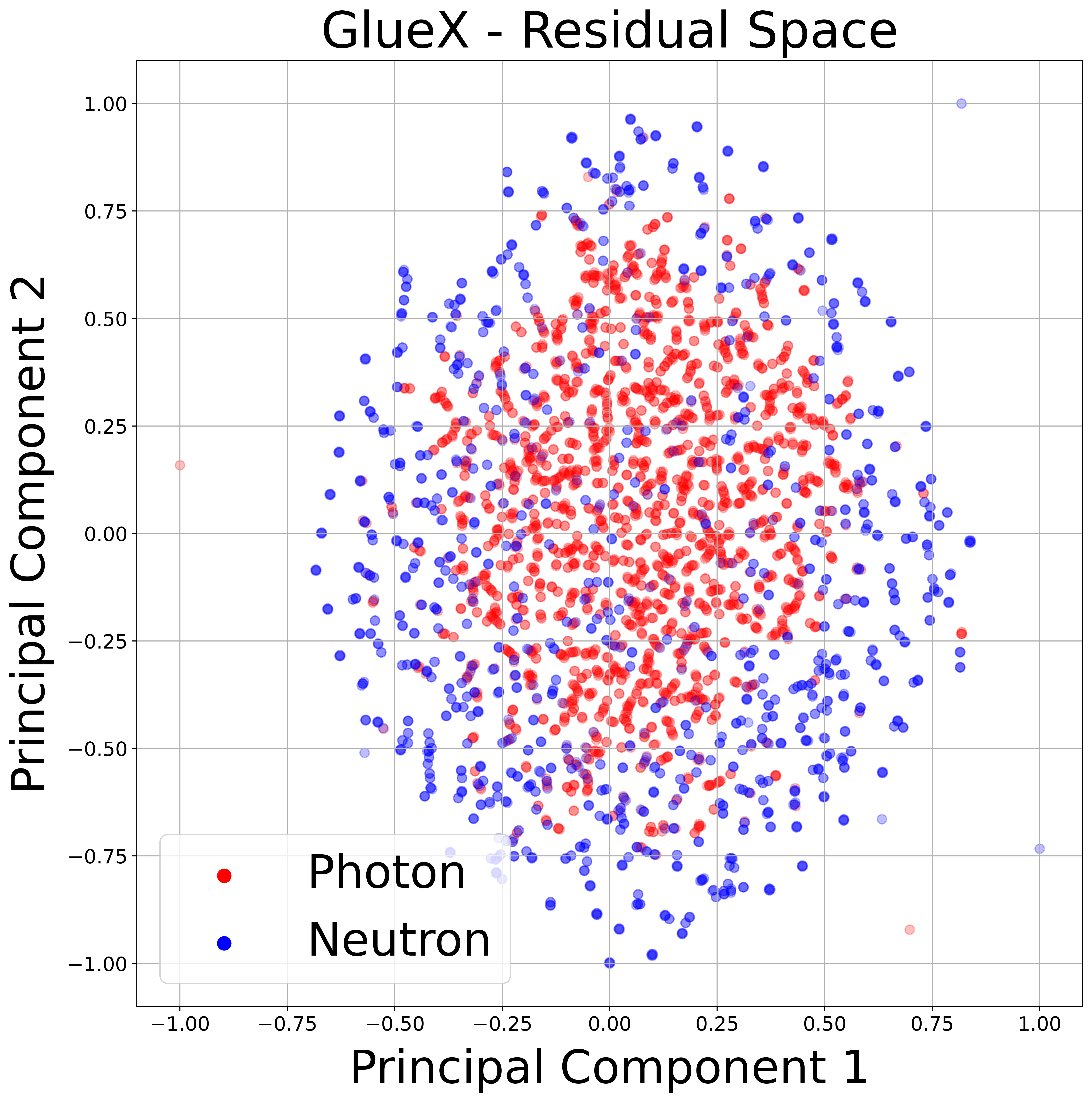} %
    \includegraphics[width=0.31\textwidth]{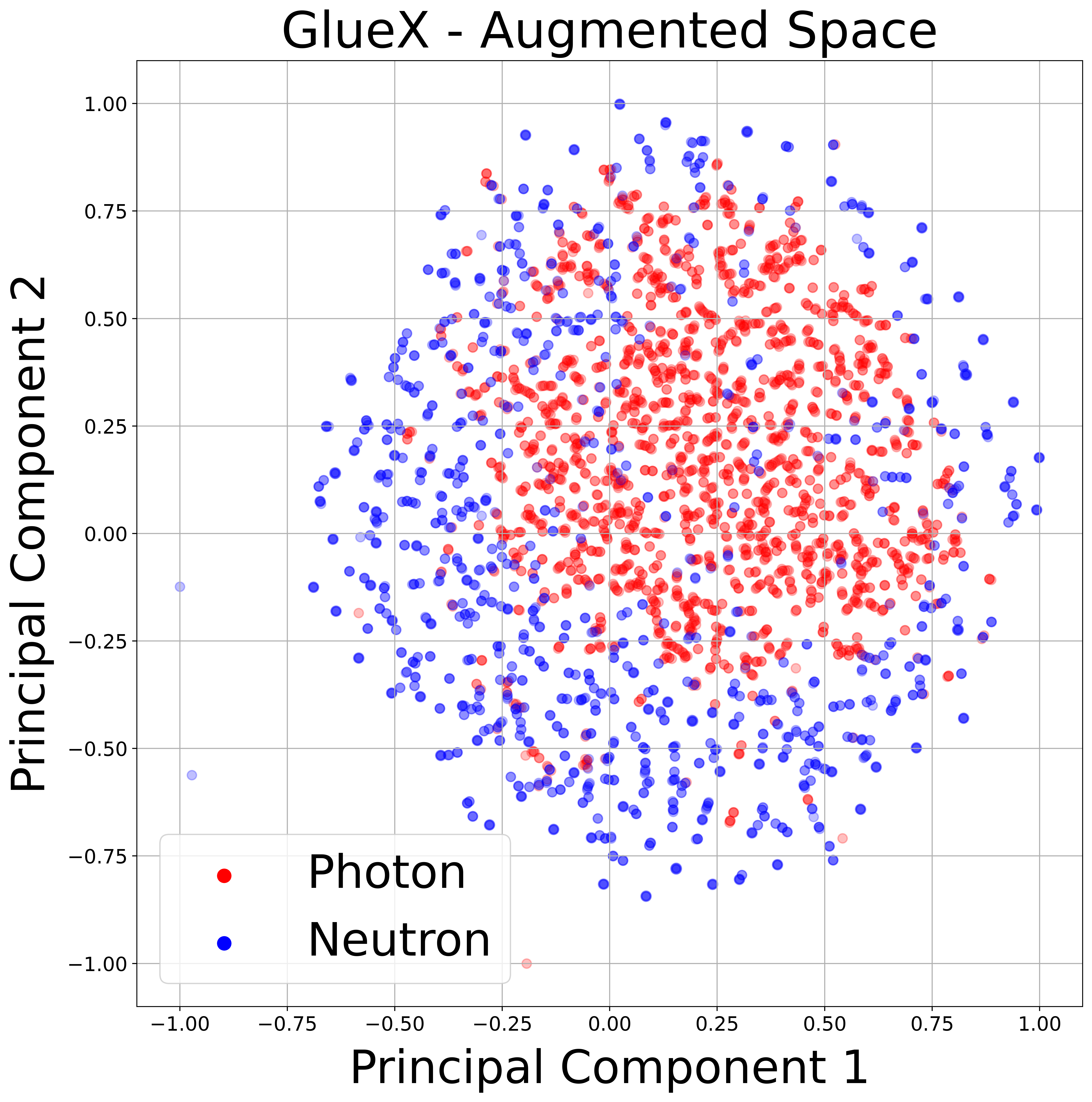}\\
    \includegraphics[width=0.31\textwidth]{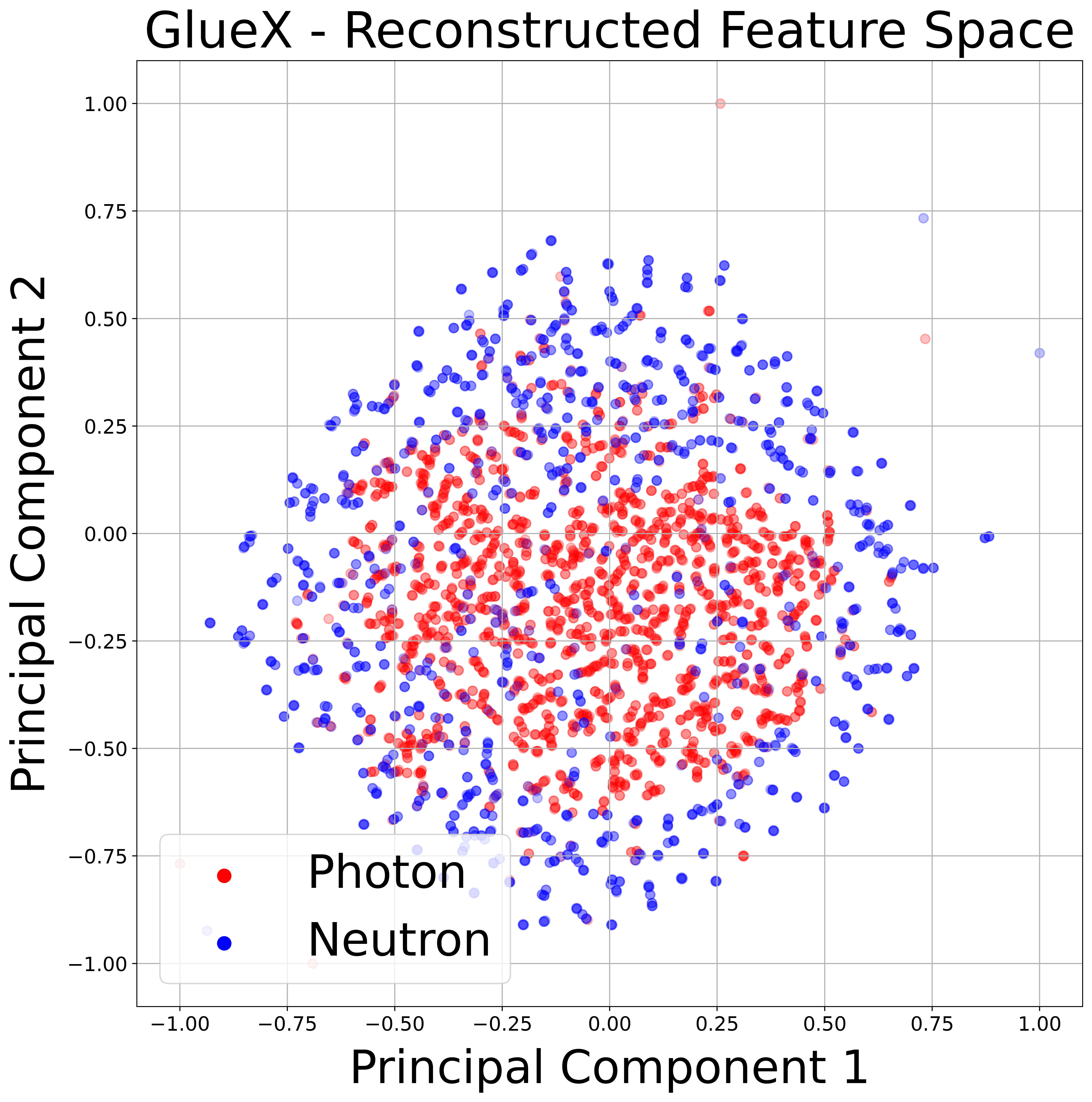}%
    \includegraphics[width=0.31\textwidth]{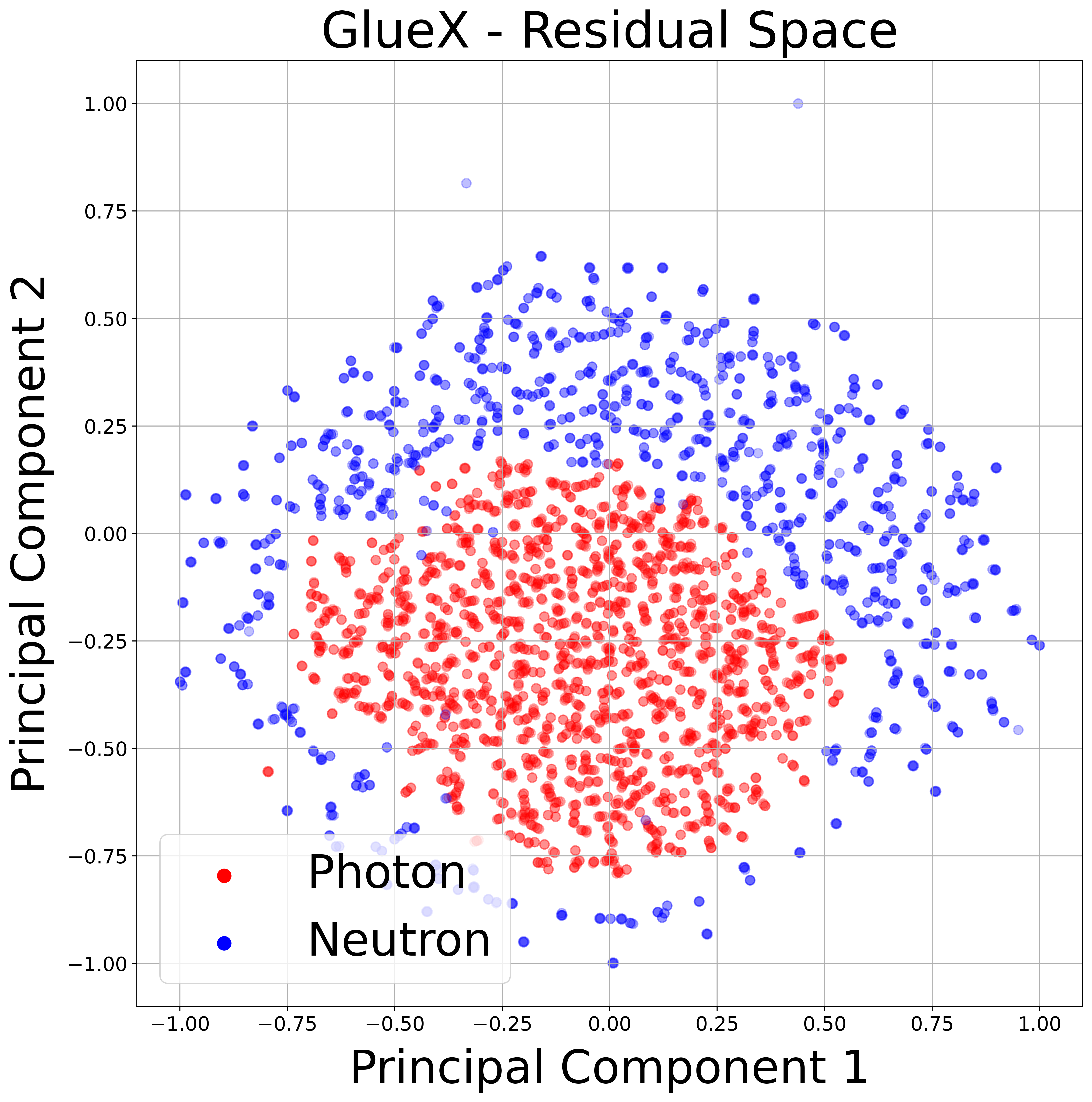}%
    \includegraphics[width=0.31\textwidth]{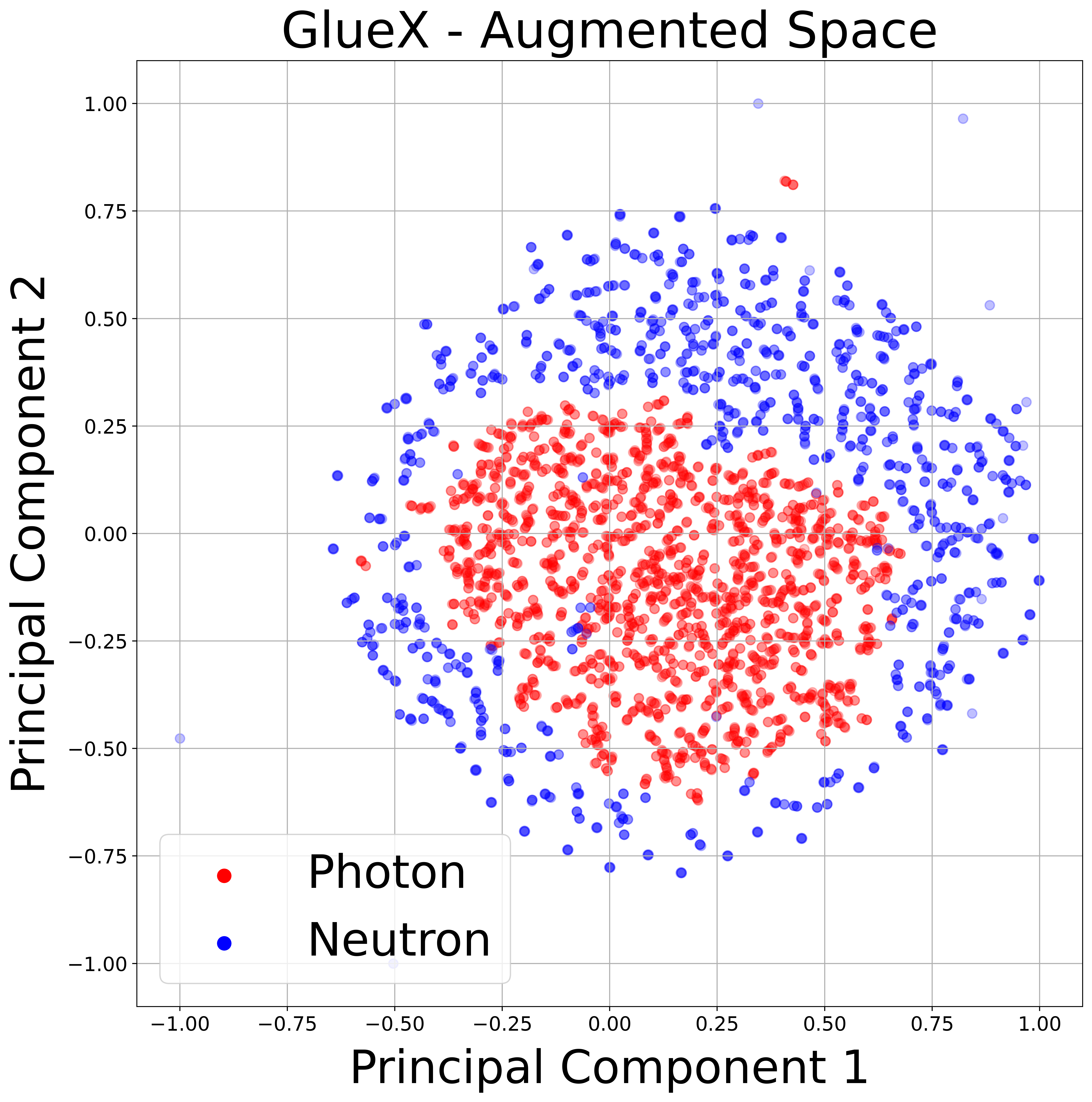}\\
    \includegraphics[width=0.31\textwidth]{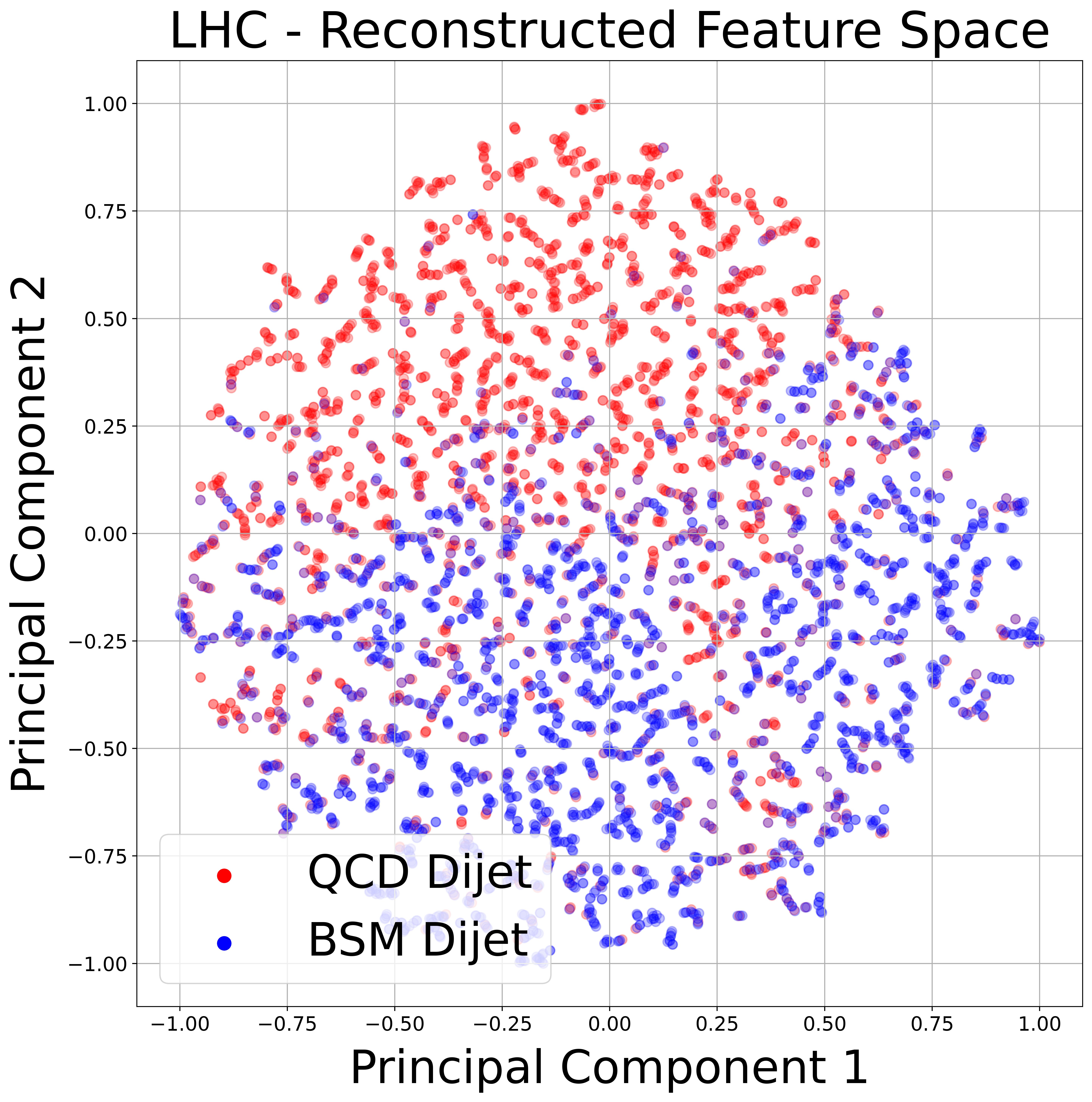}%
    \includegraphics[width=0.31\textwidth]{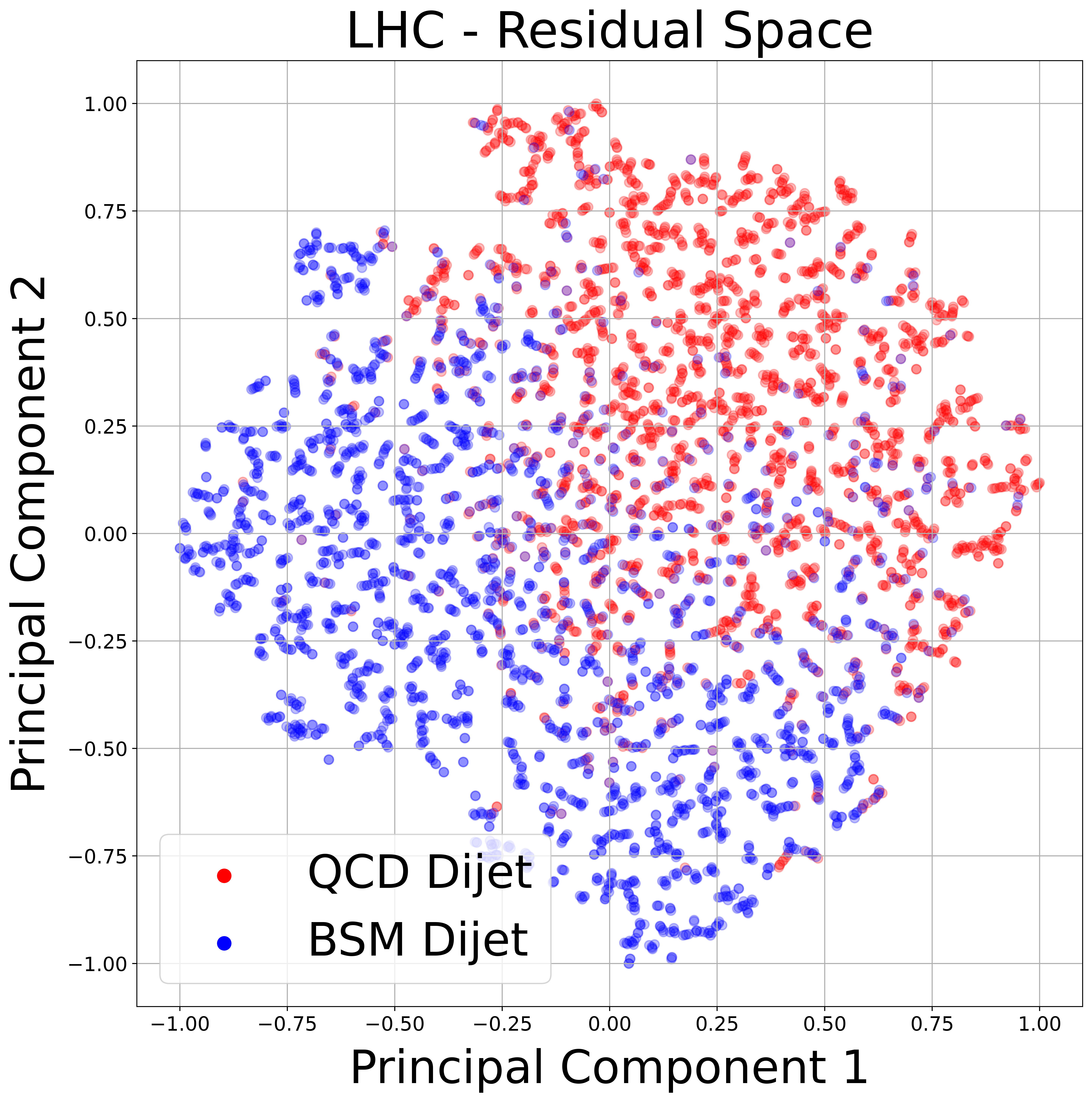}
    \includegraphics[width=0.31\textwidth]{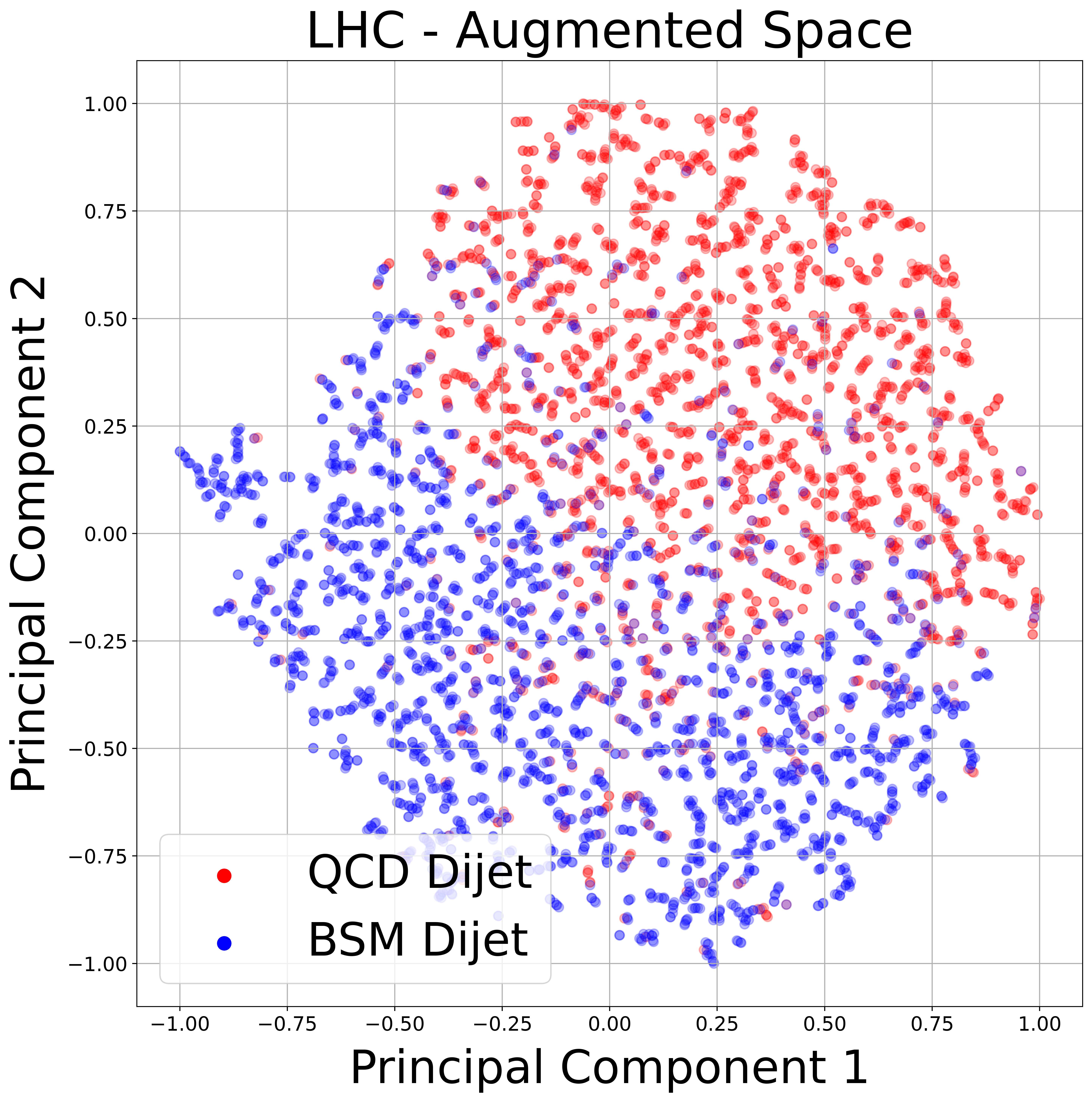}
    \caption{\textbf{Dimensionally reduced representation of the reconstructed feature, residual and augmented spaces:} t-SNE \cite{van2008visualizing} is used to provide a 2D representation of the reconstructed features, residual and augmented space. (top row) the $\gamma/n$ classification in \gluex \ BCAL (14 dimensions, 28 with augmentation); (middle row) same problem using `scaled' feature distributions for neutrons; (bottom row) the QCD dijet problem at LHC (15 dimensions, 30 with augmentation). 
    The residuals create nested clusters of higher density within the data space that are pushed radially outward from the main primary class cluster (middle column), thus allowing more accurate separation of the two classes at inference. There still exist nested clusters within the reconstructed feature space (left column), yet it is apparent that these are not as easily extracted, which explains the performance increase via the inclusion of residuals in the augmented space (right column).%
    }
    \label{fig:t-SNE}
\end{figure}

We deem the residuals to be highly valuable in terms of separation and provide further evidence that features localize the space, and the residuals push nested clusters radially outward to be more easily extracted and seen as outliers.
This interpretation is also represented in Fig. \ref{fig:t-SNE} which shows a t-SNE representation of the feature and augmented space for the \gluex \ problem. 
As shown in the figure, augmenting the space with residuals produces a better distinguishing power between photons and neutrons.  
 
\begin{table}[h]
\centering
\begin{tabular}{ c c c c c c }

& & \multicolumn{2}{c}{(\gluex)} & \multicolumn{2}{c}{(LHC)} \\
 \hline
 \textbf{Space} & \textbf{Quantile} & \textbf{TPR} & \textbf{TNR} & \textbf{TPR} & \textbf{TNR} \\ 

 \hline
 \rowcolor{blue!15} Features  & $\sigma$ (68\%) & 68.06\%  & 85.61\%  & 68.44\% & 62.15\% \\ 
 \hline
 \rowcolor{yellow!15} Augmented & $\sigma$ (68\%)  & 68.28\%  & 87.44\% & 68.18\% & 93.20\% \\ 
 \hline
 \rowcolor{blue!15} Features & $2\sigma$ (95\%)  & 95.03\%  & 47.11\% & 95.14\% & 11.20\% \\ 
 \hline 
 \rowcolor{yellow!15} Augmented & $2\sigma$ (95\%)  & 95.09\% &  52.40\% & 95.15\% & 42.40\% \\ 
 \hline
 \rowcolor{blue!15} Features & $3\sigma$ (99\%)  & 98.97\% & 25.43\% & 99.06\% & 2.29\% \\ 
 \hline
 \rowcolor{yellow!15} Augmented & $3\sigma$ (99\%)  & 98.97\%  & 34.95\% & 99.03\% & 11.82\% \\ 
 \hline

\end{tabular}
 \caption{\textbf{Features vs augmented space for \gluex \ BCAL and LHC problems:}. The benefits of residuals can be concluded via comparison of TNR at equal thresholds. Note the consistency between threshold and TPR by design in which fluctuations across differing spaces are within error.}
  \label{tab:GlueX_CERN_Residuals}
\end{table}

\subsection{Architecture specifications and computing resources}

During the inference phase, in both problems, we are limited by the generation speed of the cMAF and the clustering. HDBSCAN is in fact not optimized to run on GPU and the fitting procedure dominates the computing time.\footnote{Inference time on average is \SI{3.3}{\s} on a P100 PCIe 16GB card.}
Considering these limitations, we generate only $1.5$k reference samples per particle in inference. We show in both cases the statistics are sufficiently high to run our analyses. 

\begin{table}[!]
\setlength{\tabcolsep}{28pt}
\centering
\begin{tabular}{ c  c  c } 
\hline
\textbf{description} & \textbf{symbol} & \textbf{value} \\
\hline
Huber Loss Delta & $\delta$ & 1.0 \\
cAE Latent Space Dimension & $\mathcal{Z}$ & 6 \\
cAE Learning Rate & $\lambda_{cAE}$ & $5\cdot 10^{-4}$ \\
cMAF Learning Rate & $\lambda_{cMAF}$ & $1\cdot 10^{-4}$ \\
\gluex \ cMAF Bijections & $K_{\gluex}$ & 12 \\
LHC cMAF Bijections & $K_{LHC}$ & 10 \\
HDBSCAN Minimum Cluster Size & $M_{CS}$ & 1000 \\
HDBSCAN Minimum Samples & $M_{S}$ & 100 \\
Reference Cluster Size & $N$ & 1500 \\
\hline
\end{tabular}
 \caption{\textbf{Baseline hyperparameters of the F+M architecture:} these values are not optimal, but have shown to be reliable as an initial starting point. Thorough optimization utilizing Bayesian processes is likely to further improve performance.  
  \label{tab:hyperparameters}
  }
\end{table}

\begin{table}[!]
\setlength{\tabcolsep}{50pt}
\centering
\begin{tabular}{ c  c } 
\hline
\textbf{specs} & \textbf{value} \\
\hline
inference time per particle & $\sim 3$s \\
inference network memory & $\sim 1$GB \\
training network memory & $\sim 6$GB \\
network memory on local storage & $\sim 14$MB \\
\gluex \ cAE trainable parameters & 10,172 \\
\gluex \ cMAF trainable parameters & 3,700,704\\
LHC cAE trainable parameters & 11,545 \\
LHC cMAF trainable parameters & 3,103,920 \\
\hline
\end{tabular}
 \caption{\textbf{Specs of the F+M architecture:} the table reports the average inference time per particle, the inference memory and the training memory, \textit{i.e.}, the GPU memory required by the network during inference and training phases. Training was done utilizing google services on Nvidia A100-SMX4-40GB and Nvidia V100-16GB cards with TensorFlow 2.8.0 and TensorFlow-Probability 0.16.0 builds. Inference was supported via Compute Canada on Nvidia P100 PCIe 16GB card.
 The number of parameters correspond to the \gluex \ BCAL problem. The \gluex \ $\gamma / n$ problem has slightly larger numbers due to extra layers in the cMAF needed to reproduce multi-modal distributions in the photon sample.  
  \label{tab:specs}
  }
\end{table}

The architecture has not undergone a rigorous optimization and we expect better results could be obtained. Table~\ref{tab:hyperparameters} contains hyperparameter settings used throughout our experiments. One can imagine optimizing the pipeline end-to-end under a Bayesian process, in which we rely on the signal class only. Key hyperparameters to tune are those of the clusterer. %
 Training was done utilizing google services on Nvidia A100-SMX4-40GB and Nvidia V100-SMX2-16GB cards with TensorFlow 2.8.0 and TensorFlow-Probability 0.16.0 builds. Inference was supported via Compute Canada on Nvidia P100 PCIe 16GB card.\footnote{Cedar GPU clusters were used. See \url{https://www.computecanada.ca/} for more details.}
 The number of parameters correspond to the \gluex \ BCAL problem.
 Additional technical details on the architecture can be found in Table \ref{tab:specs}.

\section{Summary and conclusions}\label{sec:summary}

We have developed a novel architecture that consists of two steps called ``flux'' and ``mutability'' (F+M).
In the first stage we learn the distribution of a reference class. In the second stage we address if data significantly deviates from the reference class. 
The backbone of this architecture consists of: (i) a conditional autoencoder that allows to reconstruct the injected features and calculate the residuals which combined to the first make an augmented space; (ii) a conditional Masked Autoregressive Flow, that combined with a kernel density estimation allows in the inference phase to dynamically generate the reference class in the augmented space; (iii) a hierarchical-based clustering algorithm that allows to estimate if an object belongs to the reference class by producing an outlier score, which can be also seen as an estimate of how confident we are about the object belonging to the reference class.

We demonstrated its capability as a one-class-classification method when dealing with isolated neutral showers at \gluex \ BCAL, providing good separation between photons (reference class) and neutrons (complementary class) while relying on only information related to the reference class.
We then proved the advantage of this approach which is agnostic to the neutron sample, in particular when the latter is affected by large uncertainty. %

We also showed the capability of the algorithm as an anomaly detection method, isolating possible BSM $Z^{\prime} \rightarrow t\bar{t}$ dijets topologies from SM QCD dijet background (reference class) at the LHC. We demonstrated that our model performs on par with other architectures using the same dataset, yet we are able to remain truly agnostic towards the complementary class using our final quantile metric. %
A possible extension of this problem that has been left for future exploration consists in conditioning on both p$_{T}$ and the invariant mass of the leading jet.
We also note in general that our architecture has not undergone rigorous optimization of the hyperparameters and we therefore expect further increase in performance in doing so.

In both cases, we have demonstrated an increased performance via inclusion of the residuals. We have concluded that an augmented space (features + residuals) is ideal for inference. The features localize the space for a given kinematics, and the residuals push the complementary class radially outward in the hypersphere used for clustering. This allows nested clusters existing in the data space to be more easily extracted and increase distinguishing power.

The inference time per particle is fairly slow. As such, the architecture is best suited for offline analysis purposes in its current state, in which it can be optimized via parallelization. Reduction in inference time can potentially be obtained via the use of a different flow network as MAF is generally known for its slower generation speeds. Other NF models have not yet been tested and are left as further exploration. A large bulk of the inference time remains at the clustering stage as HDBSCAN is not optimized to run on GPU. Other clustering methods, or an improved GPU-optimized HDBSCAN build could potentially solve this issue. %
In the future we plan to extend this work to an application for data quality monitoring in an experiment, in that significant deviations from the expected quantiles of the reference distributions could determine if a new calibration/alignment is needed.


\section*{Acknowledgments}

%
The work of CF is supported by the U.S. Department of Energy, Office of Science, Office of Nuclear Physics under grant No. DE-SC0019999.
The work of JG and ZP is supported by the Natural Sciences and Engineering Research Council of Canada grant SAPPJ-2018-00021.
We would like to acknowledge the \gluex \  collaboration for the simulations used in the neutron identification problem.
We also want to thank E. Cisbani and M. Williams for useful comments.

\section*{References}
\bibliographystyle{iopart-num}
\bibliography{biblio}

\providecommand{\newblock}{}
\begin{thebibliography}{10}
\expandafter\ifx\csname url\endcsname\relax
  \def\url#1{{\tt #1}}\fi
\expandafter\ifx\csname urlprefix\endcsname\relax\def\urlprefix{URL }\fi
\providecommand{\eprint}[2][]{\url{#2}}

\bibitem{seliya2021literature}
Seliya N, Abdollah~Zadeh A and Khoshgoftaar T~M {A literature review on
  one-class classification and its potential applications in big data} 2021
  {\em J. Big Data\/} \href{http://dx.doi.org/10.1186/s40537-021-00514-x}{{\bf
  8} 1--31}

\bibitem{nachman2020anomaly}
Nachman B {\em {Anomaly Detection for Physics Analysis and Less Than Supervised
  Learning}\/} Artificial Intelligence for High Energy Physics
  \href{http://dx.doi.org/10.1142/9789811234033_0004}{Chapter~4, pp 85--112}

\bibitem{Kasieczka_2021}
Kasieczka G, Nachman B, Shih D, Amram O, Andreassen A {\em et~al.\/} {The {LHC}
  Olympics 2020 a community challenge for anomaly detection in high energy
  physics} 2021 {\em Rep. Prog. Phys.\/}
  \href{http://dx.doi.org/10.1088/1361-6633/ac36b9}{{\bf 84} 124201}

\bibitem{nachman2020anomaly_density}
Nachman B and Shih D {Anomaly detection with density estimation} 2020 {\em
  Phys. Rev. D\/} \href{http://dx.doi.org/10.1103/PhysRevD.101.075042}{{\bf
  101} 075042}

\bibitem{schwartz_2021}
Fraser K, Homiller S, Mishra R~K, Ostdiek B and Schwartz M~D 2021 {Challenges
  for Unsupervised Anomaly Detection in Particle Physics}
  (arXiv:\href{https://arxiv.org/abs/2110.06948}{{\tt 2110.06948}})

\bibitem{mcinnes2017hdbscan}
McInnes L, Healy J and Astels S {HDBSCAN: Hierarchical density based
  clustering} 2017 {\em J. Open Source Softw.\/}
  \href{http://dx.doi.org/10.21105/joss.00205}{{\bf 2}}

\bibitem{gluex_2021}
Adhikari S, Akondi C, Al~Ghoul H, Ali A, Amaryan M {\em et~al.\/} (GlueX
  Collaboration) {The GlueX beamline and detector} 2021 {\em Nucl. Instrum.
  Methods Phys. Res. A: Accel. Spectrom. Detect. Assoc. Equip.\/}
  \href{http://dx.doi.org/10.1016/j.nima.2020.164807}{{\bf 987} 164807} ISSN
  0168-9002

\bibitem{maf}
Papamakarios G, Pavlakou T and Murray I 2018 {Masked Autoregressive Flow for
  Density Estimation} (arXiv:\href{https://arxiv.org/abs/1705.07057}{{\tt
  1705.07057}})

\bibitem{hdbscan_imp}
McInnes L and Healy J 2017 {Accelerated Hierarchical Density Based Clustering}
  2017 IEEE Int. Conf. Data Min. Workshops ICDMW (IEEE)
  \href{http://dx.doi.org/10.1109/icdmw.2017.12}{pp 33--42}

\bibitem{campello2015hierarchical}
Campello R~J~G~B, Moulavi D, Zimek A and Sander J {Hierarchical Density
  Estimates for Data Clustering, Visualization, and Outlier Detection} 2015
  {\em ACM Trans Knowl Discov Data\/}
  \href{http://dx.doi.org/10.1145/2733381}{{\bf 10} 1--51}

\bibitem{park2021quasi}
Park S~E, Rankin D, Udrescu S~M, Yunus M and Harris P {Quasi anomalous
  knowledge: searching for new physics with embedded knowledge} 2021 {\em J.
  High Energy Phys.\/} \href{http://dx.doi.org/10.1007/JHEP06(2021)030}{{\bf
  2021} 1--26}

\bibitem{meyer2015hybrid}
Meyer C~A and Swanson E {Hybrid mesons} 2015 {\em Prog. in Part. and Nucl.
  Phys.\/} \href{http://dx.doi.org/10.1016/j.ppnp.2015.03.001}{{\bf 82} 21--58}
  (arXiv:\href{https://arxiv.org/abs/1502.07276}{{\tt 1502.07276}})

\bibitem{beattie2018construction}
Beattie T, Foda A, Henschel C, Katsaganis S, Krueger S {\em et~al.\/}
  {Construction and performance of the barrel electromagnetic calorimeter for
  the GlueX experiment} 2018 {\em Nucl. Instrum. Methods Phys. Res. A: Accel.
  Spectrom. Detect. Assoc. Equip.\/}
  \href{http://dx.doi.org/10.1016/j.nima.2018.04.006}{{\bf 896} 24--42}

\bibitem{AGOSTINELLI2003}
Agostinelli S, Allison J, Amako K, Apostolakis J, Araujo H {\em et~al.\/}
  {Geant4—a simulation toolkit} 2003 {\em Nucl. Instrum. Methods Phys. Res.
  A: Accel. Spectrom. Detect. Assoc. Equip.\/}
  \href{http://dx.doi.org/https://doi.org/10.1016/S0168-9002(03)01368-8}{{\bf
  506} 250--303} ISSN 0168-9002

\bibitem{Amram_2021}
Amram O and Suarez C~M {Tag N’ Train: a technique to train improved
  classifiers on unlabeled data} 2021 {\em J. High Energy Phys.\/}
  \href{http://dx.doi.org/10.1007/jhep01(2021)153}{{\bf 2021}} ISSN 1029-8479

\bibitem{BSM_Data}
Cheng T 2021 {Test Sets for Jet Anomaly Detection at the LHC}
  \urlprefix\url{https://doi.org/10.5281/zenodo.4614656}

\bibitem{cheng2021variational}
Cheng T, Arguin J~F, Leissner-Martin J, Pilette J and Golling T 2021
  {Variational Autoencoders for Anomalous Jet Tagging}
  (arXiv:\href{https://arxiv.org/abs/2007.01850}{{\tt 2007.01850}})

\bibitem{madgraph}
Alwall J, Frederix R, Frixione S, Hirschi V, Maltoni F {\em et~al.\/} {The
  automated computation of tree-level and next-to-leading order differential
  cross sections, and their matching to parton shower simulations} 2014 {\em J.
  High Energy Phys.\/} \href{http://dx.doi.org/10.1007/jhep07(2014)079}{{\bf
  2014}} ISSN 1029-8479

\bibitem{pythia}
Sjöstrand T, Ask S, Christiansen J~R, Corke R, Desai N, Ilten P, Mrenna S,
  Prestel S, Rasmussen C~O and Skands P~Z {An introduction to PYTHIA 8.2} 2015
  {\em Comput. Phys. Commun.\/}
  \href{http://dx.doi.org/10.1016/j.cpc.2015.01.024}{{\bf 191} 159–177} ISSN
  0010-4655

\bibitem{delphes}
de~Favereau J, Delaere C, Demin P, Giammanco A, Lemaître V {\em et~al.\/}
  {DELPHES 3: a modular framework for fast simulation of a generic collider
  experiment} 2014 {\em J. High Energy Phys.\/}
  \href{http://dx.doi.org/10.1007/jhep02(2014)057}{{\bf 2014}} ISSN 1029-8479

\bibitem{SM_Data}
Leissner-Martin J, Cheng T and Arguin J~F 2020 {QCD Jet Samples with Particle
  Flow Constituents} \urlprefix\url{https://doi.org/10.5281/zenodo.4641460}

\bibitem{fastjet}
Cacciari M, Salam G~P and Soyez G {FastJet user manual} 2012 {\em Eur. Phys. J.
  C\/} \href{http://dx.doi.org/10.1140/epjc/s10052-012-1896-2}{{\bf 72}} ISSN
  1434-6052

\bibitem{anti_kt}
Cacciari M and Salam G~P {Dispelling the N$^{3}$ myth for the $k_{t}$
  jet-finder} 2006 {\em Phys. Lett. B\/}
  \href{http://dx.doi.org/10.1016/j.physletb.2006.08.037}{{\bf 641} 57–61}
  ISSN 0370-2693

\bibitem{Kasieczka_2020}
Kasieczka G, Marzani S, Soyez G and Stagnitto G {Towards machine learning
  analytics for jet substructure} 2020 {\em Journal of High Energy Physics\/}
  \href{http://dx.doi.org/10.1007/jhep09(2020)195}{{\bf 2020}}

\bibitem{beattie_2022}
Beattie T 2022 private communication

\bibitem{xgb}
Chen T and Guestrin C 2016 {XGBoost: A Scalable Tree Boosting System} Proc.
  22nd ACM SIGKDD Int. Conf. Knowl. Discov. Data Min. (ACM)
  \href{http://dx.doi.org/10.1145/2939672.2939785}{p 785–794} ISBN
  9781450342322

\bibitem{van2008visualizing}
Van~der Maaten L and Hinton G {Visualizing data using t-SNE.} {\em J. Mach.
  Learn. Res.\/} {\bf 9} 2579--2605
  \url{http://jmlr.org/papers/v9/vandermaaten08a.html}

\end{thebibliography}

\clearpage
\appendix

\section{\gluex \ BCAL feature definitions} \label{app:features}

All showers are \textbf{DBCALShower} objects in the \gluex \ software package. We denote these showers as $S$, and label the quantity we use via a subscript ($S_x$ denotes the x position of the shower object for example). $R$ denotes the inner radius of the BCAL (\SI{64.3}{\cm} radially outwards from the center line of the target) and $T_z$ denotes the center z position of the hydrogen target (\SI{65}{\cm} from the upstream edge of the BCAL). Showers are comprised of points, we define our features using the energy weighting of these points.\footnote{Hit - SiPM trigger at one end of the BCAL. Point - SiPM trigger at both ends of the BCAL. Points are used for: a) to find $z$ position of a shower. b) noise reduction (two ended coincidence).} 
%

\begin{itemize}
\item\textbf{LayerM\_E} $ = \sum_i^N E_i$ 
\\ $M \in \{1,2,3,4\}$ is the layer number and $E_i$ is the energy of the $i^{th}$ reconstructed point in the layer. \\
\item \textbf{Layer$M$bySumLayers\_E} $= \frac{1}{E_{total}} \sum_i^N E_i$
\\ $M \in \{1,2,3,4\}$ is the layer number and $E_i$ is the energy of the $i^{th}$ reconstructed point in the layer. \\
\item \textbf{Z Width} $= \sqrt{\frac{1}{E_{total}}\sum_i^N E_i(\Delta z_i)^2}$ ,\;  $\Delta z_i$ = ($z_i$ + $T_z$) - $S_z$
\\$E_i$ and $z_i$ are the energy and z position of the $i^{th}$ point in the shower. \\
\item \textbf{R Width} $= \sqrt{\frac{1}{E_{total}}\sum_i^N E_i(\Delta r_i)^2}$ ,\; $\Delta r_i$ = ($R$ - $r_i$)
\\$E_i$ and $r_i$ are energy and radial position of the $i^{th}$ point.\\
\item \textbf{T Width} $= \sqrt{\frac{1}{E_{total}}\sum_i^N E_i(\Delta t_i)^2}$ ,\; $\Delta t_i$ = $t_i - S_t$
\\$E_i$ and $t_i$ are the energy and timing information of the $i^{th}$ point.\\
\item $\boldsymbol{\theta}$ \textbf{Width} $= \sqrt{\frac{1}{E_{total}}\sum_i^N E_i(\Delta \theta_i)^2}$ ,\; $\Delta \theta_i$ = $\theta_i - S_{\theta}$
\\$E_i$ and $\theta_i$ are the energy and polar angle (from the target center) of the $i^{th}$ point.\\
\item $\boldsymbol{\phi}$ \textbf{Width} $= \sqrt{\frac{1}{E_{total}}\sum_i^N E_i(\Delta \phi_i)^2}$ ,\; $\Delta \phi_i$ = $\phi_i - S_{\phi}$
\\$E_i$ and $\phi_i$ are the energy and azimuthal angle of the $i^{th}$ point.\\
\item $\boldsymbol{z}$ \textbf{Entry} = $(S_z - T_z)\dfrac{R}{S_r} + T_z$
\\The position at which the particle hits the inner radius of the BCAL.
\end{itemize}
\section{\gluex \ kinematic plots}\label{app:gluex_kinematics}

\begin{figure}[b]
    \centering
    \includegraphics[trim=0 0cm 0 0, width=1.\textwidth]{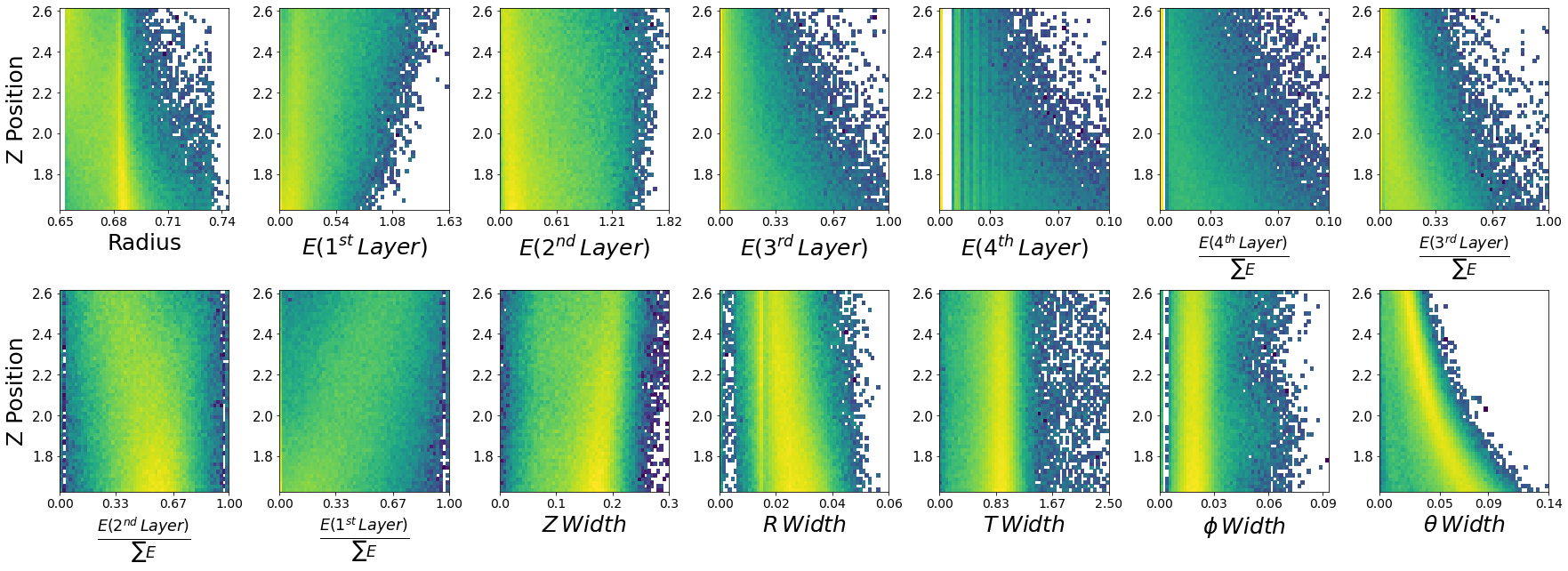}
    \includegraphics[trim=0 0cm 0 0, width=1.\textwidth]{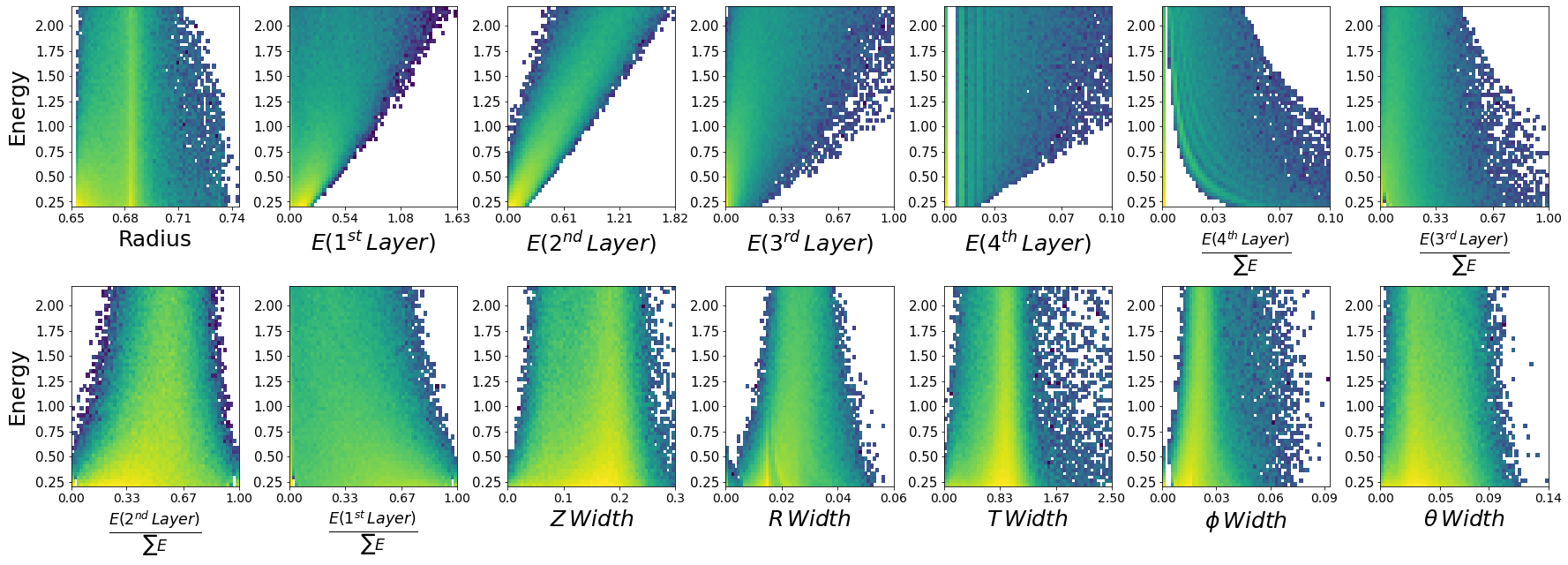}
    \caption{\textbf{2D histograms of photon features (x-axis) and z position (or energy) (y-axis):} We see a clear dependence on $z$ and energy for most of the features within the space. (top two rows) Features such as T Width, and $\phi$ Width display less of a z dependence on average. 
    (bottom two rows) Any feature corresponding to energy deposition within layers has a large dependence on $z$ and $E$, along with width variables.
    By conditioning on $z$ and $E$, we are able to capture the functional dependence of detector response at the generation stage. %
    }
    \label{fig:photons_f_z}
\end{figure}

\begin{figure}[!]
    \centering
    \includegraphics[trim=0 0cm 0 0, width=1.\textwidth]{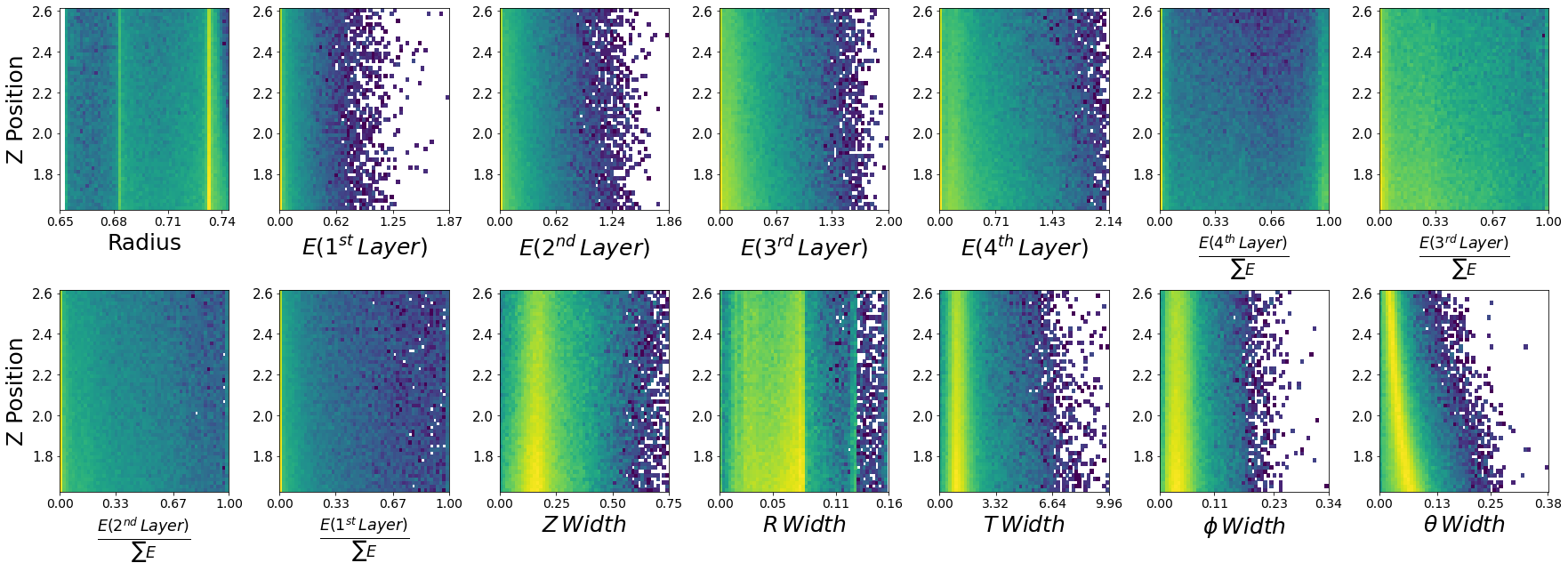}
    \includegraphics[trim=0 0cm 0 0, width=1.\textwidth]{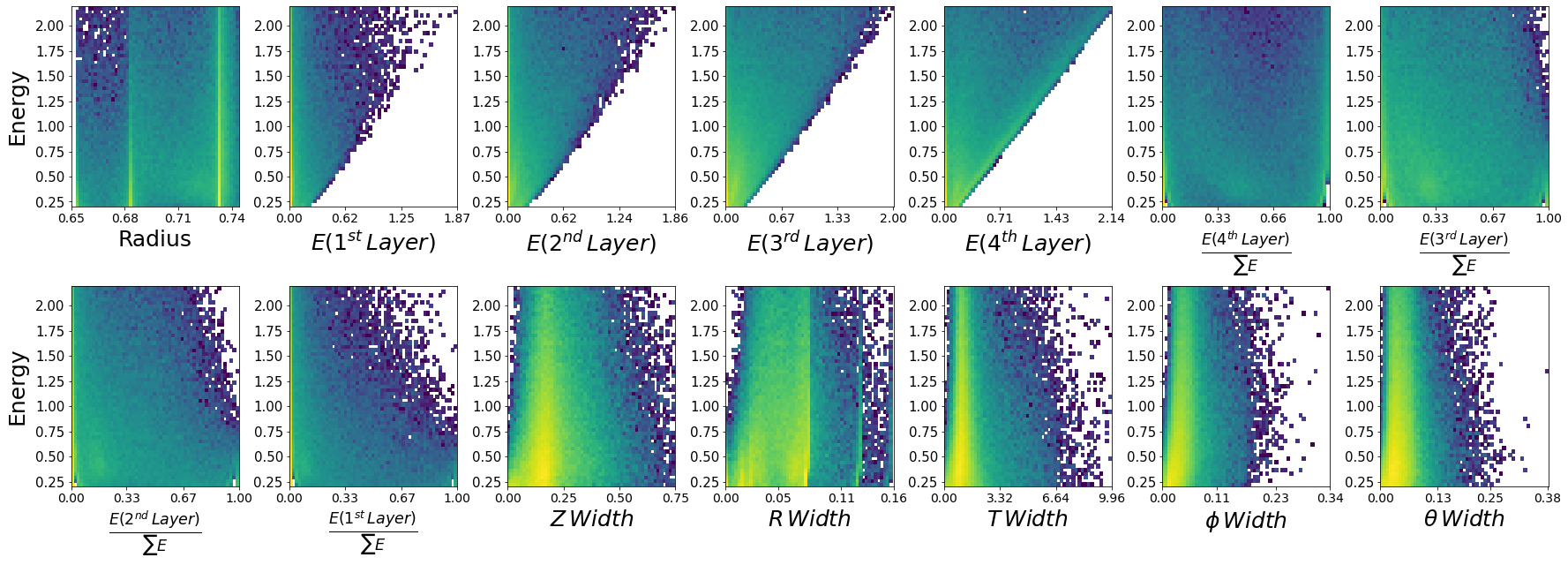}
    \caption{\textbf{2D histograms of neutron features (x-axis) and z position (or energy) (y-axis):} (top two rows) We see a lesser degree of z position dependence on neutron features in comparison to that of photons. Features with high dependence in the photon sample no longer exhibit the degree of functional relation in the neutron sample due to their differing interactions. 
    (bottom two rows) We see a clear dependence on energy within the feature space: any feature corresponding to energy deposition within layers has a high dependence, along with width variables.
    The dependence differs from that of photons, but overlaps at certain regions in the phase space making separation of the two classes more difficult.
    }
    \label{fig:neutrons_f(z)}
\end{figure}

Plots contained in this section illustrate the 14 features (see~\ref{app:features}) as a function of kinematics for both simulated photons and neutrons, as described in the captions. These plots are integrated over the entire phase space used in the analysis. The functional dependence of features on the phase space ($E$,$z$) can be clearly seen in the simulated samples, more so for the photons as the simulation of a neutron interaction is more difficult. As one moves to training on real data from standard ``candles'' such as $\pi^0 \rightarrow \gamma \gamma$, the dependencies become more pronounced. 
The distributions of the features with respect to the $z$ position and energy in BCAL are shown in Fig.~\ref{fig:photons_f_z} and \ref{fig:neutrons_f(z)} for photons and neutrons, respectively.

\section{\gluex \ generations}\label{app:gluex_generations}

\begin{figure}[!]
    \centering
    \includegraphics[width=\textwidth]{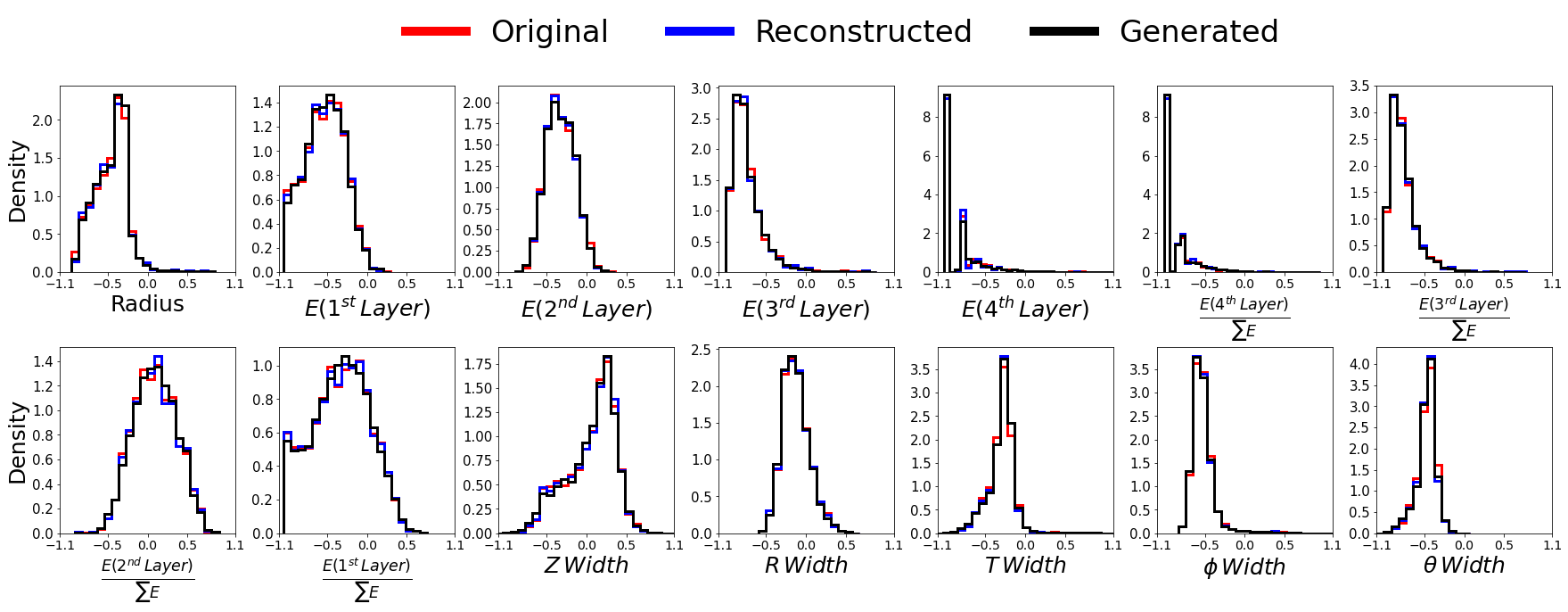} \\
    \includegraphics[width=\textwidth]{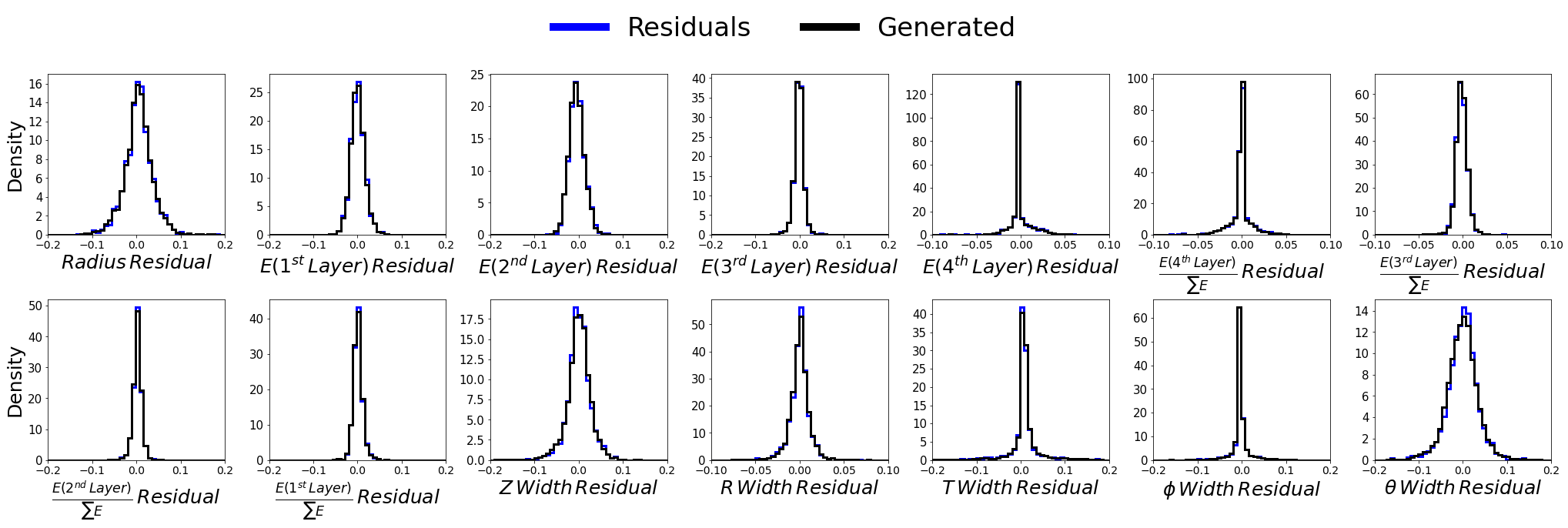}
    \caption{\textbf{Features of reconstructed photon showers in \gluex \ BCAL:} Original, reconstructed and generated features (top two rows), residuals and generated residuals (bottom two rows), for \SI{1}{\GeV} $< E <$ \SI{1.4}{\GeV}, \SI{206}{\cm} $< z <$ \SI{218}{\cm}. The cMAF is trained to generate the reconstructed space of the cAE as it was found to give better separation power due to skewing of OOD samples (\textit{i.e.} neutrons). The cMAF matches closely the distributions of the sampled, reconstructed photons.
    }
    \label{fig:gluex_generations}
\end{figure}

We conditionally generate data using the cMAF in a select kinematic region to demonstrate the quality of data produced. A central region of the of the phase space we are working with (\SI{1}{\GeV} $< E <$ \SI{1.4}{\GeV}, \SI{206}{\cm} $ < z <$ \SI{218}{\cm}) is chosen, and used to compare three different quantities, namely the original injected features, reconstructions from the cAE and the generations of the cMAF. For each original point within the phase space, we generate ten artificial data points, as such, generated distributions an order of magnitude larger in terms of sample size but have been normalized. Being that we are generating the reconstructed space of the cAE, found to be beneficial due to skewing of out-of-distribution (OOD) samples, one may argue the use of a Conditional Variational Autoencoder (cVAE) may be appropriate. We have developed a similar algorithm using cVAE's although it is not optimal for a few reasons, namely, the quality of generations and also the problem with dead nodes (referring to vanishing gradients, in which outputs tend to zero) during training. Due to small values in the residual space, the training process can be very tricky and can lead to Dirac delta distributions at zero in some variables if dead nodes occur. Using an NF instead was found to be more reliable and it can be seen from Fig.~\ref{fig:gluex_generations} the distributions are consistent.

\begin{figure}[!]
    \centering
    \includegraphics[trim=0 0cm 0 0, width=1.\textwidth]{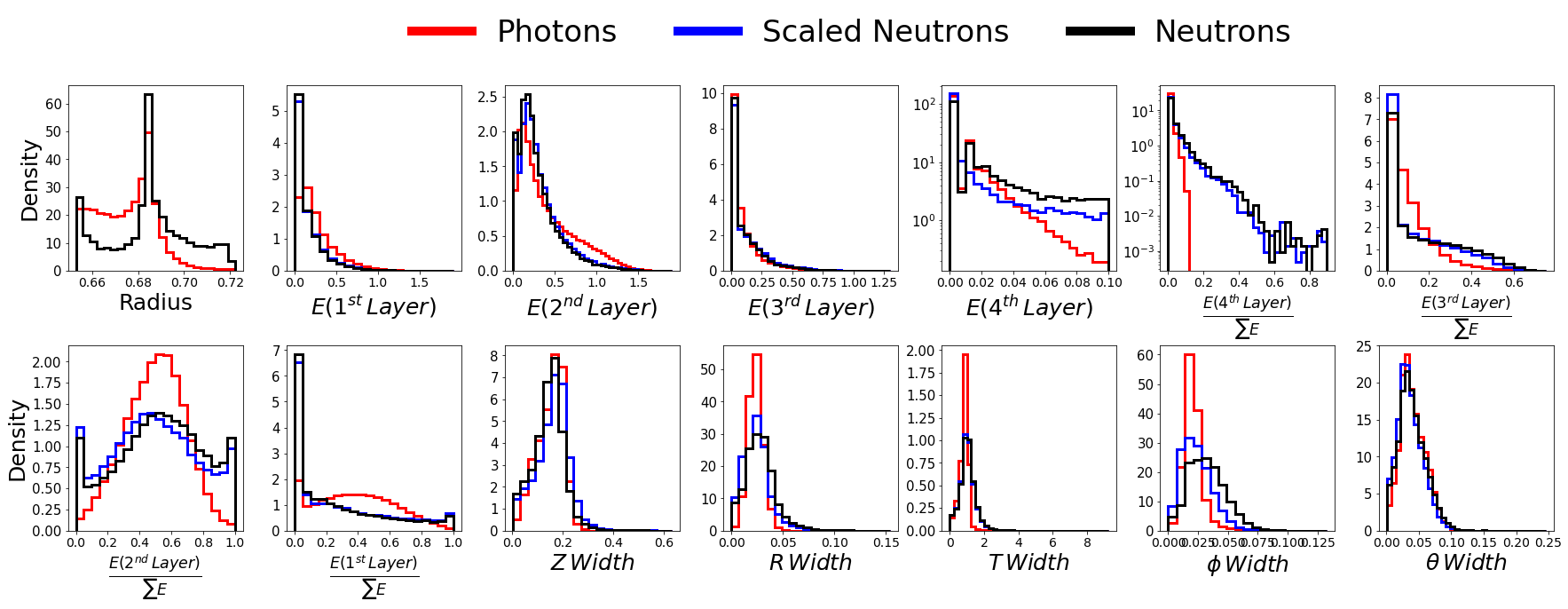}
    \caption{\textbf{Photon and neutrons distributions:} Photon and neutron distributions. Original and scaled neutron distributions are also shown for comparison.}
    \label{fig:pert_original}
\end{figure}

\begin{figure}[!]
    \centering
    \includegraphics[trim=0 0cm 0 0, width=1.\textwidth]{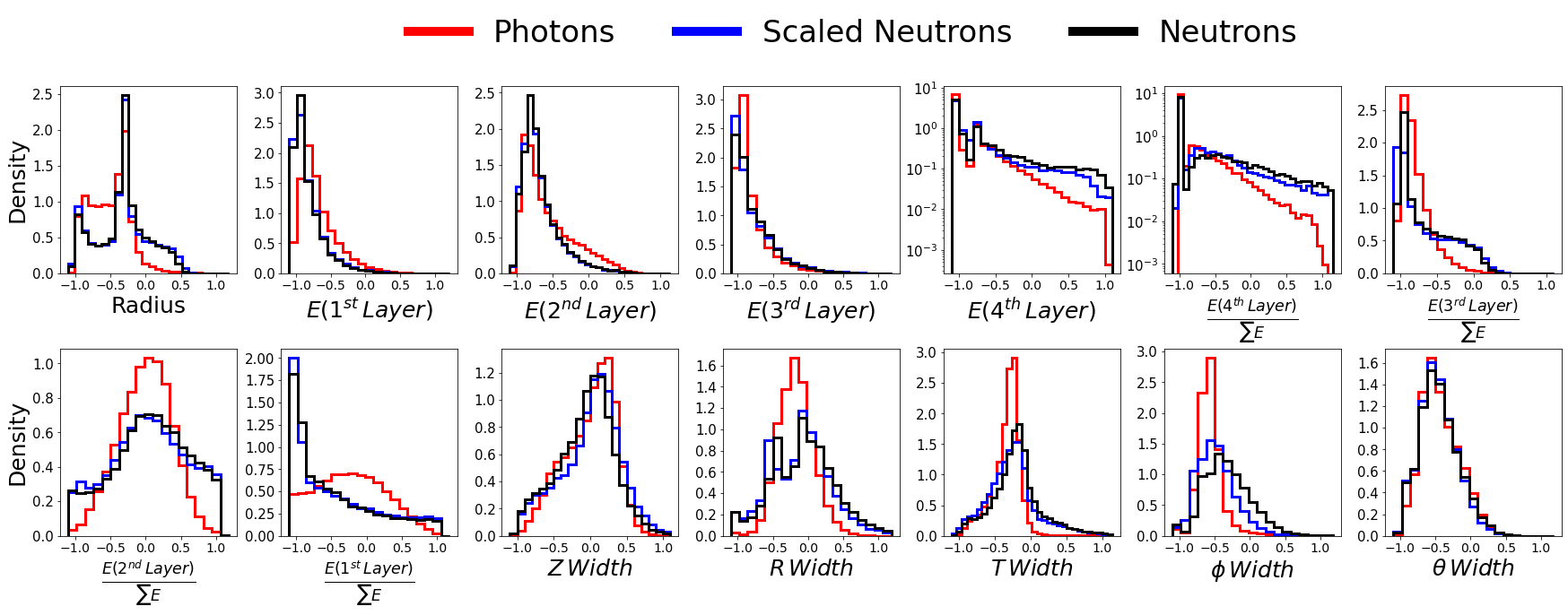}
    \caption{\textbf{cAE reconstructed photon and neutrons distributions:} Photon and neutron distributions for the features reconstructed by the cAE. Original and scaled neutron distributions are also shown for comparison.}
    \label{fig:pert_recon}
\end{figure}

\begin{figure}[!]
    \centering
    \includegraphics[trim=0 0cm 0 0, width=1.\textwidth]{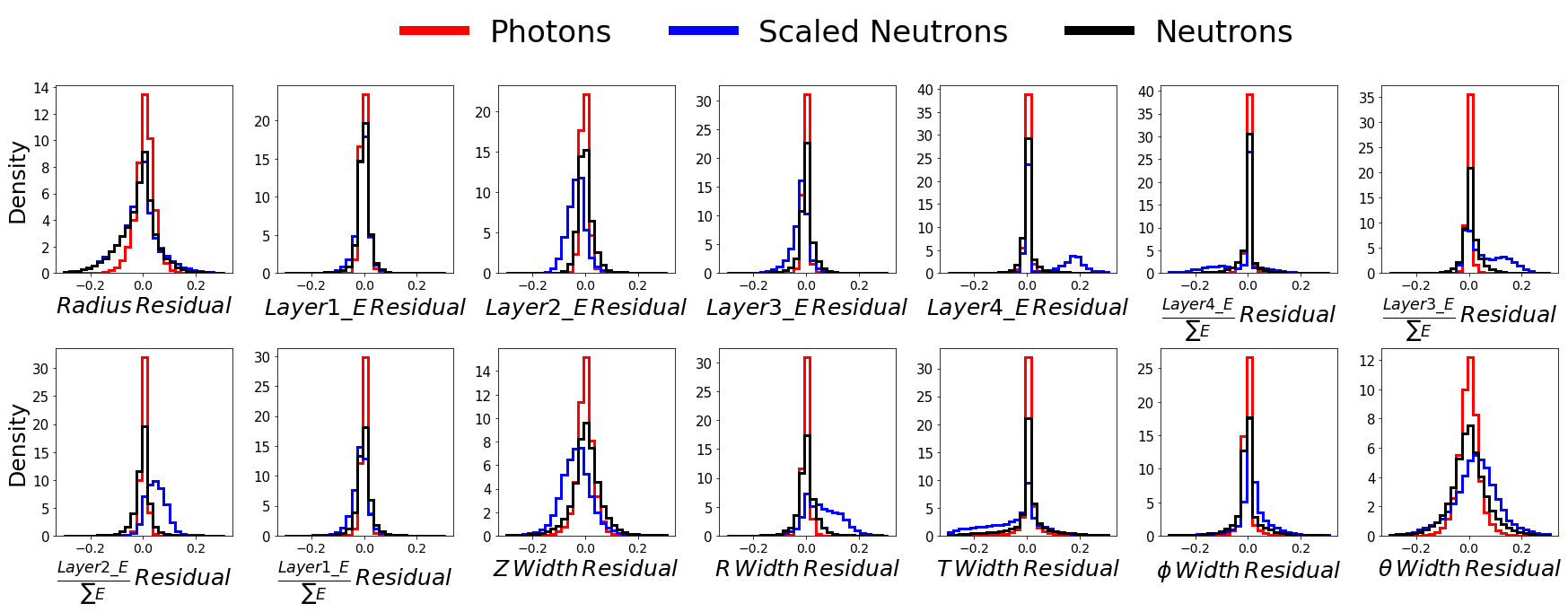}
    \caption{\textbf{cAE residuals for photon and neutrons distributions:}  Photon and neutron distributions for the residuals obtained with the cAE. Original and scaled neutron distributions are also shown for comparison.}
    \label{fig:pert_residuals}
\end{figure}

\begin{figure}[!]
    \centering
    \includegraphics[trim=0 0cm 0 0, width=.5\textwidth]{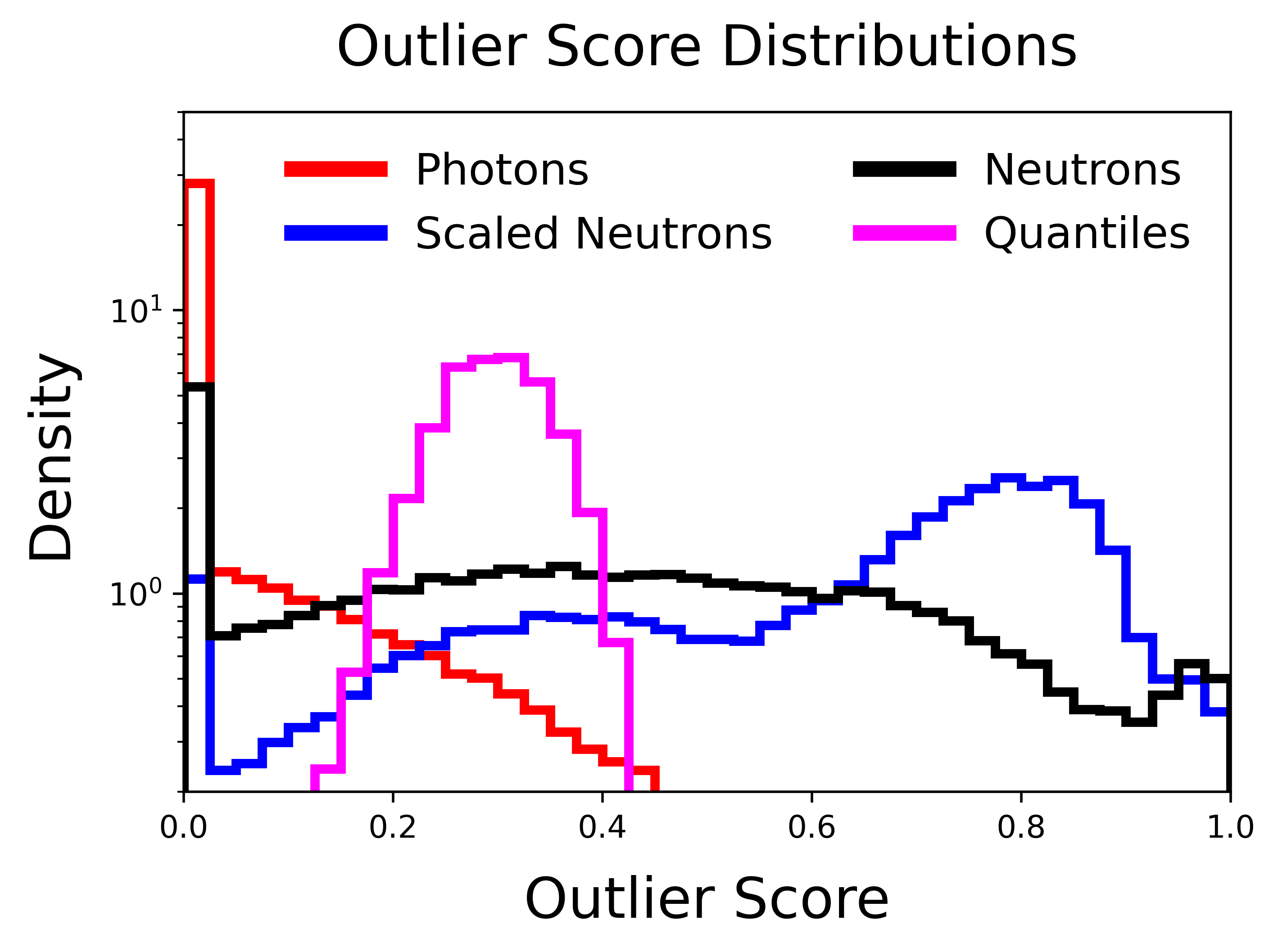}
    \caption{\textbf{Outlier score distributions:} Photon, neutron, perturbed neutron and quantile cut (coinciding with the TPR of XGBoost equal to 92.15\%) distributions. The perturbed neutron outlier scores are on average higher than the original, given that the cAE is able to detect kinematic discrepancies introduced via the perturbation. %
    }
    \label{fig:perturbation_outlier_scores}
\end{figure}

On a particle-by-particle basis we scaled the neutron features through this empirical formula: $x +S*P*\frac{(x-x_{min})*(x_{max}-x)}{(x_{max}-x_{min})^2}$, where $S$ being the sign that controls which direction to nudge, $P$ is the scaling effect in percentage (we used a 10\% effect, that is $P$=90\% or 110\% depending on the sign $S$), and $x_{min}$, $x_{max}$ are the minimum and maximum in the feature range which are physically allowed on each feature. We applied this scaling to all features with the exception of the shower $R$ for which it has been neglected.
Fig.~\ref{fig:pert_original} shows the injected distributions for photons, neutrons and scaled neutrons; Fig.~\ref{fig:pert_recon} shows the corresponding reconstructed features, whereas Fig. \ref{fig:pert_residuals} shows the residuals; Fig.~\ref{fig:perturbation_outlier_scores} shows the corresponding outlier score distributions obtained with F+M.

\vspace{30cm}
\section{LHC generations}\label{app:lhc_generations}

We conditionally generate data using the cMAF in a select kinematic region to demonstrate the quality of data produced. A central region of the of the phase space we are working with (600 $GeV$ $< p_{T_{j1}} <$ 650 $GeV$) is chosen, and used to compare three different quantities, namely the original features, reconstructions from the AE and the generations of the cMAF. For each original point within the phase space, we generate ten artificial data points, as such, generated distributions an order of magnitude larger in terms of sample size but have been normalized. 
\begin{figure}
    \centering
    \includegraphics[width=\textwidth]{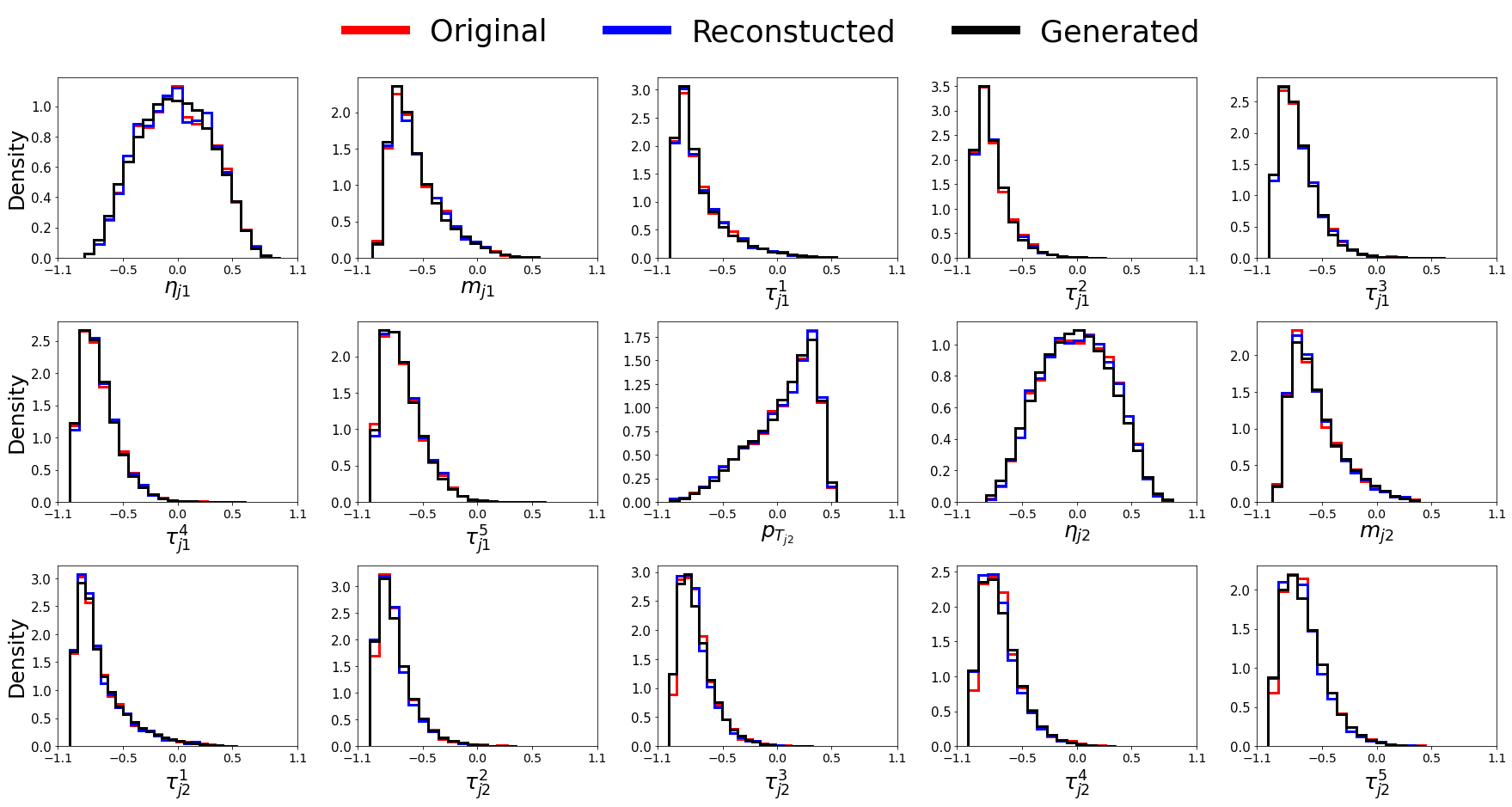} \\
    \includegraphics[width=\textwidth]{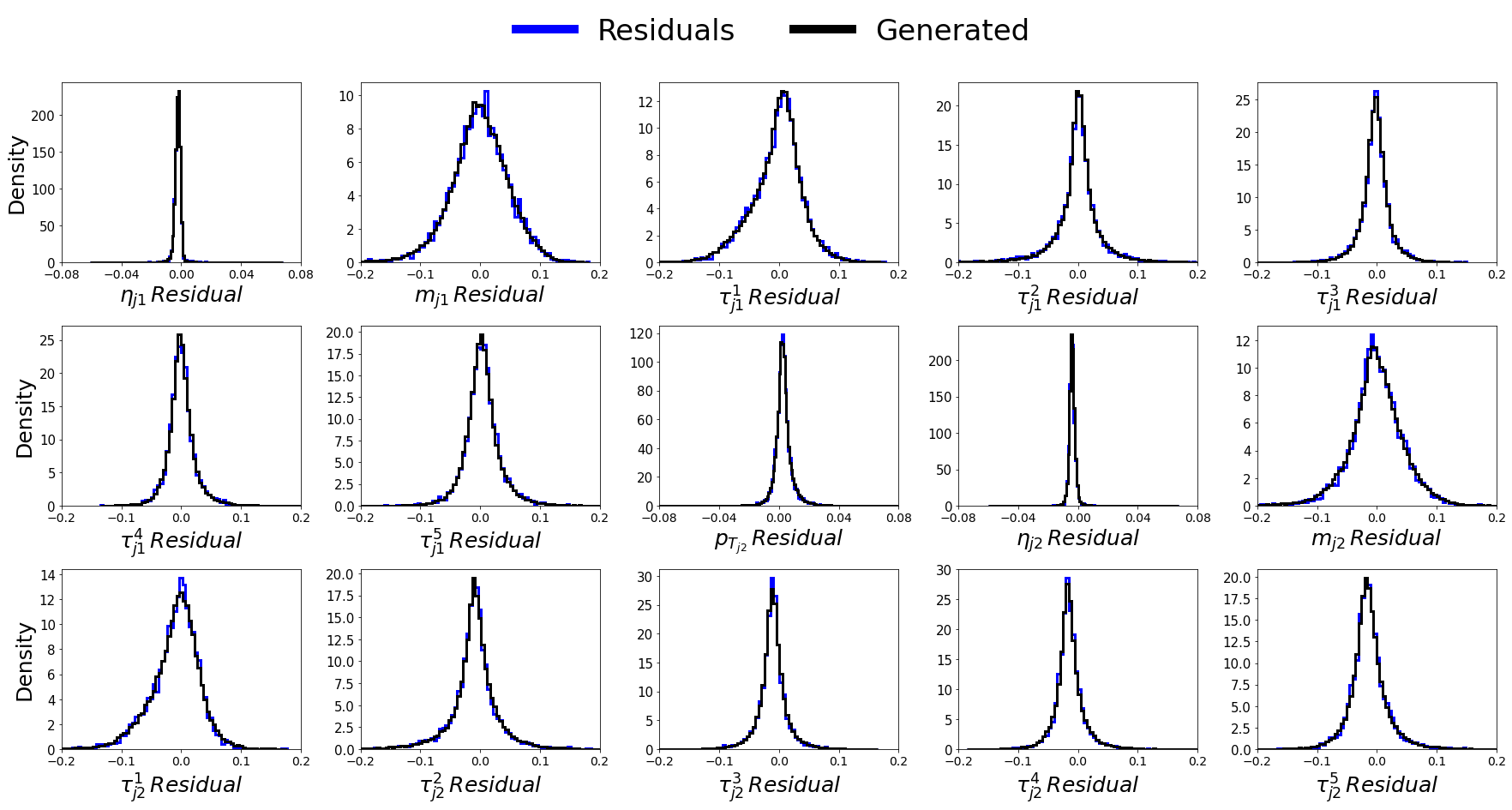}
    \caption{\textbf{Features of QCD dijet events at LHC:} Original, reconstructed and generated features (top three rows), residuals and generated residuals (bottom three rows), for 600 $GeV$ $< p_{T_{j1}} <$ 650 $GeV$. The cMAF is trained to generate the reconstructed feature space of the cAE as it was found to give better separation power due to skewing of OOD samples (i.e. Top Jets). The cMAF matches the distributions of the reconstructed QCD dijets to a very high degree. }
    \label{fig:lhc_generations}
\end{figure}
\begin{figure}
    \centering
    \includegraphics[width=\textwidth]{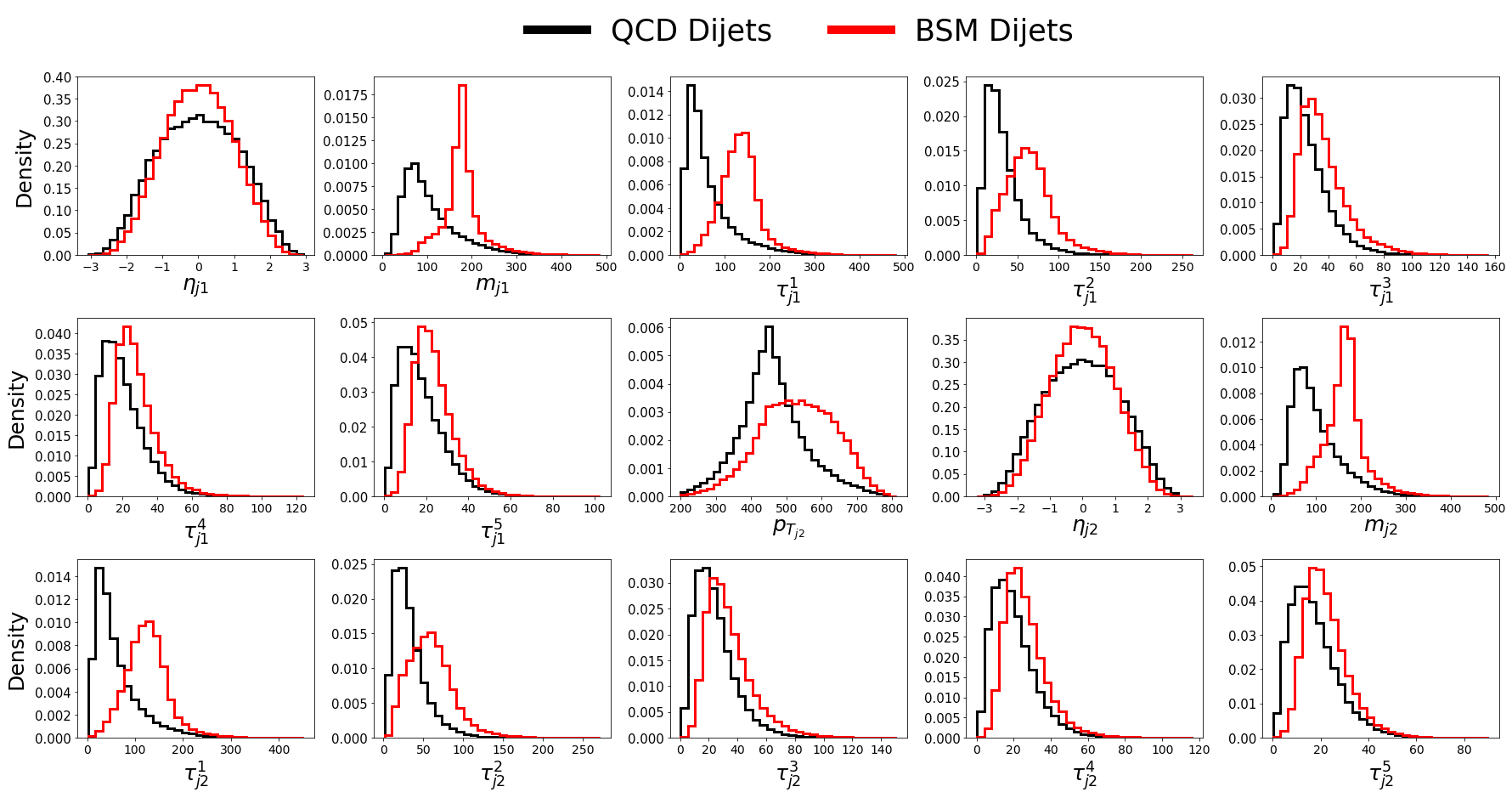}
    \caption{\textbf{Features of QCD dijet and BSM dijet events at LHC:} Feature distributions are integrated over the entire phase space. The features overlap to a high degree, yet the resulting means of distributions occupy different regions within the space. The resulting differences in kinematic correlations are able to be exploited via the residuals.}
    \label{fig:lhc_features}
\end{figure}
We notice in this dataset the generation quality is not as good as for \gluex \ in~\ref{app:gluex_generations}, this is due to the relatively low training sample size and resulting large kinematic bins (\SI{1}{\GeV} in transverse momenta) in the KDE modeling phase. The training data set should ideally be more dense (in terms of continuous conditionals) which would allow smaller modeling with KDE, and overall a more robust learning phase for the cMAF. We can see that when training data is sufficient, the generations are extremely accurate (see~\ref{app:gluex_generations}). The discrepancies in some variables undoubtedly affect performance although it can still be seen the clusterer is able to efficiently use injected data at the inference phase based off performance obtained.

Fig.~\ref{fig:lhc_generations} shows the distributions of the SM QCD dijet events at LHC, for the injected, reconstructed and generated data; Fig.~\ref{fig:lhc_features} shows a comparison between the SM QCD and the BSM feature distributions.

\end{document}